\documentclass[9pt,twocolumn,twoside]{osajnl}

\usepackage{graphicx}
\usepackage{tabularx}
\usepackage{multirow}
\usepackage{arydshln}
\usepackage{wrapfig}
\usepackage{subcaption}

\usepackage[utf8]{inputenc} 
\usepackage[T1]{fontenc}    
\usepackage{hyperref}       
\usepackage{booktabs}       
\usepackage{amsfonts}       
\usepackage{nicefrac}       
\usepackage{microtype}      
\usepackage{xcolor}         
\usepackage{amsmath}
\usepackage{bm}
\usepackage{makecell}

\newcommand{\figref}[1]{Fig.~\ref{#1}}
\newcommand{\figreftwo}[4]{Figs.~\ref{#1}#2 and~\ref{#3}#4}
\newcommand{\eqnref}[1]{Eq.~\ref{#1}}
\newcommand{\tabref}[1]{Table~\ref{#1}}

\journal{jocn} 

\setboolean{shortarticle}{false}

\title{Fully neuromorphic vision and control\\ for autonomous drone flight} 

\author[ ]{F. Paredes-Vall\'es$^{\dagger}$}
\author[ ]{J.\,J. Hagenaars$^{\dagger\,\ast}$}
\author[ ]{J.\,D. Dupeyroux$^{\dagger}$}
\author[ ]{S. Stroobants}
\author[ ]{Y. Xu}
\author[ ]{G.\,C.\,H.\,E. de Croon}

\affil[ ]{Micro Air Vehicle Laboratory, Faculty of Aerospace Engineering, Delft University of Technology, Delft, Netherlands}
\affil[*]{To whom correspondence should be addressed: j.j.hagenaars@tudelft.nl.}
\affil[$\dagger$]{Equal contribution.}

\dates{ }

\doi{}  
\setboolean{displaycopyright}{false}

\begin{abstract}
Biological sensing and processing is asynchronous and sparse, leading to low-latency and energy-efficient perception and action. In robotics, neuromorphic hardware for event-based vision and spiking neural networks promises to exhibit similar characteristics. However, robotic implementations have been limited to basic tasks with low-dimensional sensory inputs and motor actions due to the restricted network size in current embedded neuromorphic processors and the difficulties of training spiking neural networks. Here, we present the first fully neuromorphic vision-to-control pipeline for controlling a freely flying drone. Specifically, we train a spiking neural network that accepts high-dimensional raw event-based camera data and outputs low-level control actions for performing autonomous vision-based flight. The vision part of the network, consisting of five layers and 28.8k neurons, maps incoming raw events to ego-motion estimates and is trained with self-supervised learning on real event data. The control part consists of a single decoding layer and is learned with an evolutionary algorithm in a drone simulator. Robotic experiments show a successful sim-to-real transfer of the fully learned neuromorphic pipeline. The drone can accurately follow different ego-motion setpoints, allowing for hovering, landing, and maneuvering sideways - even while yawing at the same time. The neuromorphic pipeline runs on board on Intel's Loihi neuromorphic processor with an execution frequency of 200~Hz, spending only 27~\textmu J per inference. These results illustrate the potential of neuromorphic sensing and processing for enabling smaller, more intelligent robots.
\end{abstract}

\begin{document}

\maketitle

\section{Introduction}
Over the past decade, deep artificial neural networks (ANNs) have revolutionized the field of artificial intelligence. Among the successes has been the significant improvement of visual processing, to an extent that computer vision can now outperform humans on specific tasks \cite{ciresan2011committee}. Also the field of robotics has benefited from this development, with deep ANNs achieving state-of-the-art performance in tasks such as stereo vision \cite{cheng2020hierarchicala,gu2020cascadea}, optical flow estimation \cite{ilg2017flowneta,sun2018pwcnet,teed2020rafta}, segmentation \cite{yuan2021segmentation,liu2022swina}, object detection \cite{girshick2015fasta,redmon2018yolov3a,xu2021endtoend}, and monocular depth estimation \cite{garg2016unsuperviseda,godard2017unsuperviseda,yuan2022newa}. However, this high performance typically relies on substantial neural network sizes that require quite heavy and power-hungry processing hardware. This limits the number of tasks that can be performed by large robots, such as self-driving cars, and even prevents deployment on smaller robots with highly stringent resource constraints, like small flying drones. 

Neuromorphic hardware may provide a solution to this problem, since it mimics the sparse and asynchronous nature of sensing and processing in biological brains~\cite{indiveri2000neuromorphica}. For example, the pixels in neuromorphic, event-based cameras only transmit information on brightness changes \cite{gallego2020eventbased}. Since typically only a fraction of the pixels change in brightness significantly, this leads to sparse vision inputs with subsequent events that are in the order of a microsecond apart. The asynchronous and sparse nature of visual inputs from event-based cameras represents a paradigm shift compared to traditional, frame-based computer vision. Ideally, processing would exploit these properties for quicker, more energy-efficient processing. However, currently, the main approach to event-based vision processing is to accumulate events over a substantial amount of time, creating an ``event window'' that represents extended temporal information. This window is then processed similarly to a traditional image frame with an ANN \cite{zhu2018evflownet,zhu2019unsuperviseda,gehrig2021eraft,paredes-valles2021back}. One important avenue to achieve the full potential of neuromorphic vision is to process events asynchronously as they come in by means of neuromorphic processors designed for implementing spiking neural networks (SNNs) \cite{maass1997networksa,gruning2014spikinga}. These networks have temporal dynamics more similar to biological neurons. In particular, the neurons have a membrane voltage that integrates incoming inputs and causes a spike when it exceeds a threshold. The binary nature of spikes allows for much more energy-efficient processing than the floating point arithmetic in traditional ANNs. The energy gain is further improved by reducing the spiking activity as much as possible, as is also a main property of biological brains \cite{sterling2015principlesa}. Coupling neuromorphic vision to neuromorphic processing promises low-energy and low-latency visual sensing and acting, as exhibited by agile animals such as flying insects \cite{muijres2014fliesa}.

In this article, we present the first fully neuromorphic vision-to-control pipeline for controlling a freely flying drone, demonstrating the potential of neuromorphic hardware. To achieve this, we overcome several challenges related to present-day neuromorphic sensing and processing. For example, training is currently still much more difficult for SNNs than for ANNs \cite{pfeiffer2018deepa,tavanaei2019deepa}, mostly due to their sparse, binary, and asynchronous nature. Whereas continuous values can directly serve as input or output in ANNs, they have to be encoded or decoded in SNNs. Although various coding options are available \cite{bialek1989readinga,dupeyroux2022toolboxa,schuman2022evaluatinga}, it is still far from clear what the best choices are given a problem's properties\textemdash as it is also still unknown how biological brains encode information \cite{friston1997anothera,eggermont1998therea,jazayeri2017navigatinga,guest2017what}. The most well-known difficulty of SNN learning is the non-differentiability of the spiking activation function, which prevents naive application of backpropagation. Currently, this is tackled rather successfully with the help of surrogate gradients \cite{neftci2019surrogatea,zenke2021remarkable}. Moreover, while the richer neural dynamics can potentially represent more complex temporal functions, they are also harder to shape; and neural activity may saturate or dwindle during training, preventing further learning. The causes for this are hard to analyze, as there are many parameters that can play a role. Depending on the model, the relevant parameters may range from neural leaks and thresholds to recurrent weights and time constants for synaptic traces. A solution may lie in learning these parameters \cite{chowdhury2021understanding,fang2021incorporating}, but this further increases the dimensionality of the learning problem. Finally, when targeting a robotics application, SNN training and deployment is further complicated by the restrictions of existing embedded neuromorphic processing platforms, which are typically still rather limited in terms of numbers of neurons and synapses. As an illustration, the ROLLS chip~\cite{qiao2015reconfigurable} accommodates 256 spiking neurons, the Intel Kapoho Bay (featuring two Loihi chips \cite{davies2018loihi} in a USB stick form factor) 262.1k neurons \cite{vitale2021eventdriven}, and the SpiNNaker version in \cite{galluppi2014eventbased} 768k neurons. Although these chips differ in many more aspects than only the  number of neurons, this small sample already shows that current state-of-the-art SNNs cannot be easily embedded on robots. SNNs that have recently been trained on visually complex tasks such as optical flow determination \cite{paredes-valles2020unsupervised,hagenaars2021selfsupervised},  still feature far too large network sizes for implementation on current neuromorphic processing hardware for embedded systems. The smallest size SNN in these studies is LIF-FireFlowNet for optical flow estimation \cite{hagenaars2021selfsupervised}, which still has 3.7M neurons (at an input resolution of 128x128).

As a consequence, pioneering work in this area has been limited in complexity. Very early work involved the evolution of spiking neural network connectivity to map the 16 visual brightness inputs of a wheeled Kephera robot to its two motor outputs \cite{floreano2001evolution}. The evolved SNN, simulated in software, allowed the robot to avoid the walls in a black-and-white-striped environment. Most work exploring SNNs for robotics focuses on simulation. For example, in \cite{bing2018enda}, the events from a simulated event-based camera with 128x128 pixels are accumulated into frames, compressing them over time into 8x4 Poisson input neurons. These inputs, which capture the clear white lines of the road border, are then directly mapped to two output neurons for staying in the center of the road with the help of reward-modulated spike-time-dependent plasticity (R-STDP) learning. Robotic examples of in-hardware neuromorphic processing are more rare. An early example is the one in \cite{galluppi2014eventbased}, in which an event-based camera with 128x128 pixels is connected to a SpiNNaker neuromorphic processor to allow a driving robot to differentiate between two lights flashing at different frequencies with a 128-neuron winner-takes-all network. In \cite{milde2017obstacle} a spiking neural network is designed for following a light target in the top half of the field of view, while avoiding regions with many events in the bottom half of the field of view. This network is successfully implemented in the ROLLS neuromorphic chip~\cite{qiao2015reconfigurable} and tested in an office environment. Recent years have seen an increasing focus on flying robots, i.e., drones, because they need to react quickly while being extremely restricted in terms of size, weight, and power (SWaP). In \cite{vitale2021eventdriven}, an SNN is implemented on a bench-fixed dual-rotor to align the roll angle with a black-and-white disk located in front of the camera. The SNN involved both a visual Hough transform \cite{ballard1981generalizinga} for finding the line, and a proportional-derivative (PD) controller for generating the propeller commands. Finally, in \cite{dupeyroux2021neuromorphica}, an SNN was first evolved in simulation and then implemented in Loihi for vision-based landing of a freely flying drone. This control network only consisted of 35 neurons since the visual processing was still performed with conventional, frame-based computer vision methods. Additionally, it is worth noting that only the vertical motion of the drone was controlled with the SNN; its lateral position was controlled using traditional control algorithms and an external motion capture system.

\subsection*{A fully neuromorphic solution to vision-based navigation}

\begin{figure*}[!h]
	\centering
	\includegraphics[width=\textwidth]{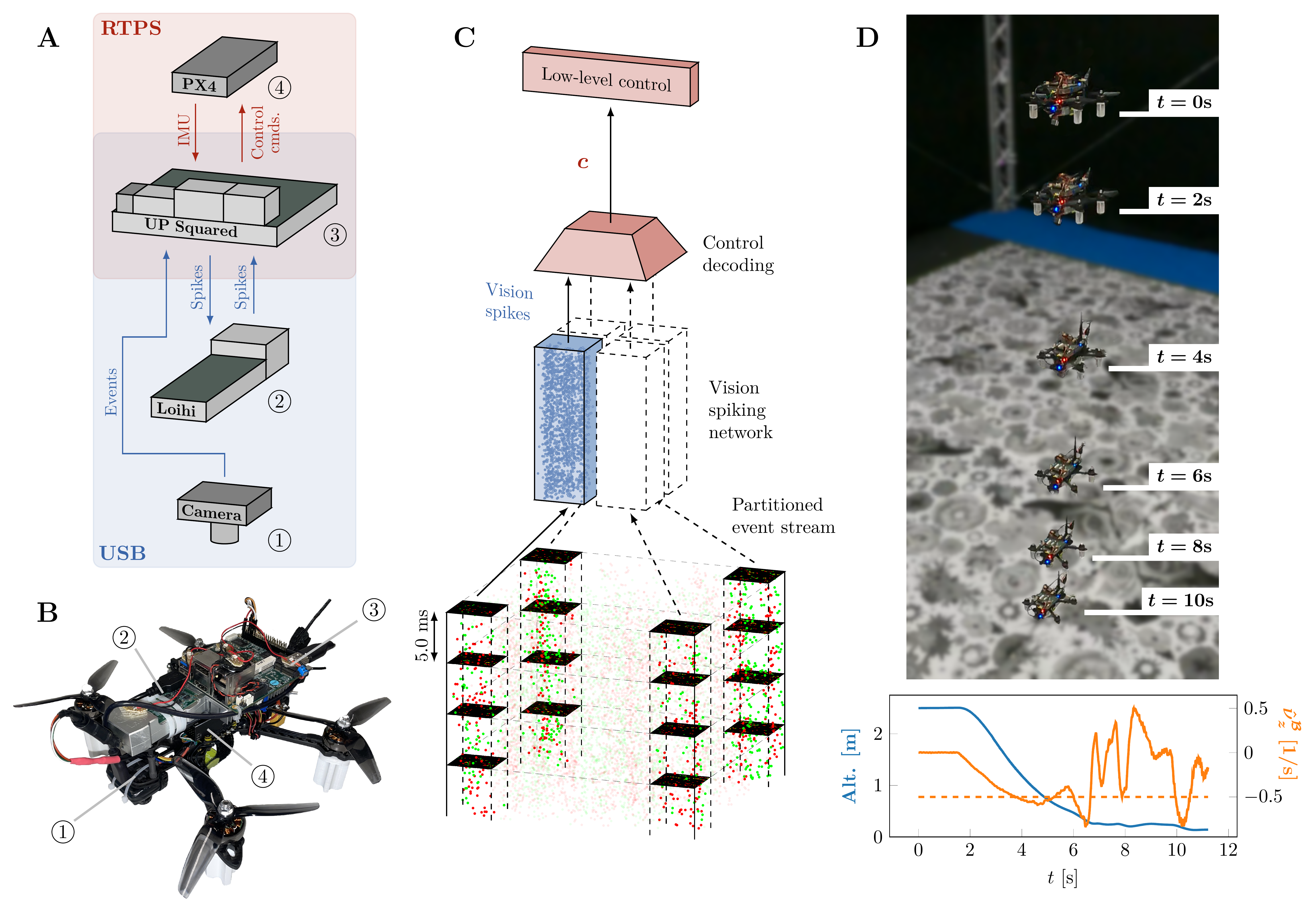}
	\caption{\textbf{Overview of the proposed system.} \textbf{(A)} Quadrotor used in this work (total weight 1.0 kg, tip-to-tip diameter 35 cm). \textbf{(B)} Hardware overview showing the communication between event-camera, neuromorphic processor, single-board computer and flight controller \textbf{(C)} Pipeline overview showing events as input, processing by the vision network and decoding into a control command. \textbf{(D)} Demonstration of the system for an optical flow divergence landing.} \label{fig:pipeline}
\end{figure*}

The presented vision-to-control pipeline consists of a spiking neural network that is trained to accept high-dimensional raw event-camera data and output low-level control actions for performing autonomous vision-based ego-motion estimation and control at approximately 200 Hz. Moreover, we propose a learning setup that evades the issue of the slow and inaccurate simulation of event-based vision inputs for control policy learning. In particular, it splits vision and control, so that the vision part of the network can be trained with self-supervised learning, and the control policy can be learned in a drone simulator that does not need to simulate events. The resulting pipeline, illustrated in \figref{fig:pipeline}C, was implemented on the Loihi neuromorphic processor~\cite{davies2018loihi} and used on board a small flying robot (see \figref{fig:pipeline}B) for vision-based navigation. A schematic of the hardware setup employed is shown in \figref{fig:pipeline}A. The system successfully follows ego-motion setpoints in a fully autonomous fashion, i.e., without any external aids such as a positioning system. \figref{fig:pipeline}D shows an example of a landing experiment with our neuromorphic pipeline in the control loop of the drone. The figure shows the smoothly decreasing height (blue line), and the optical flow divergence, which is the vertical component of the scaled velocity vector $\smash{\bm{\nu}^\mathcal{B} = \bm{v}^\mathcal{B}/p_z^\mathcal{WB}}$, where $p_z^\mathcal{WB}$ is the height of the drone above the ground. The divergence curve is typical of an optical flow divergence landing, first approaching the setpoint $\smash{\nu^\mathcal{B}_{z,sp}=-0.5}$~1/s and then becoming more oscillatory when getting very close to the ground~\cite{decroon2016monocular}.

As mentioned, the main challenge of deploying such a pipeline on embedded neuromorphic hardware is that, due to the preliminary state of this technology, one has to work within very tight limits regarding the available computational resources. In this project, several design decisions were made to adapt to these limitations. Firstly, the vision processing pipeline assumes that the event-based camera on the drone, the DAVIS240C \cite{brandli2014240a}, looks down at a static flat surface. Knowing the structure of the visual scene in advance simplifies the estimation of the ego-motion of the camera (and hence of the drone) with the help of optical flow information, as in \cite{decroon2013opticflow,decroon2016monocular,pijnackerhordijk2018vertical,hagenaars2020evolveda,dupeyroux2021neuromorphica}. Optical flow, i.e., the apparent motion of scene points in the image space, can be estimated from the output of an event-based camera with a wide variety of methods, ranging from sparse feature-tracking algorithms \cite{benosman2012asynchronousa} to dense (i.e., per-pixel) machine learning models \cite{zhu2018evflownet,gehrig2021eraft,hagenaars2021selfsupervised}. In the search for an efficient and high-bandwidth vision pipeline, the second design decision was to reduce the spatial resolution of the event-based vision data by only processing information from the image corners rather than the entire image space. More specifically, as depicted in \figreftwo{fig:pipeline}{C}{fig:vision}{A}, we propose the use of a small spiking neural network that is applied independently at each image corner, with each corner being 16x16 pixels in size after a nearest-neighbor downsampling operation. Each network consists of 7.2k neurons and 506.4k synapses distributed over five spiking layers, i.e., one input layer, three self-recurrent encoders, and a pooling layer. Its parameters (i.e., weights, thresholds, and leaks) are identical for the four corners, and it estimates the optical flow, in pixels per millisecond, of the corresponding corner. Because of the static and planar scene assumption, the apparent motion of the scene points at the four image corners encodes non-metric information about the velocity of the camera (i.e., scaled by the distance to the surface along the optical axis) and its rotational rates in a linear manner \cite{baker2006parameterizing}. We use this relation, combined with a linear control layer trained in simulation, to convert the spikes that encode the optical flow directly into thrust and attitude control commands (given a setpoint). This allows us to tackle the vision-based control of a freely flying drone in a fully neuromorphic fashion.

\begin{figure*}[!t]
    \centering
    \includegraphics[width=0.975\textwidth]{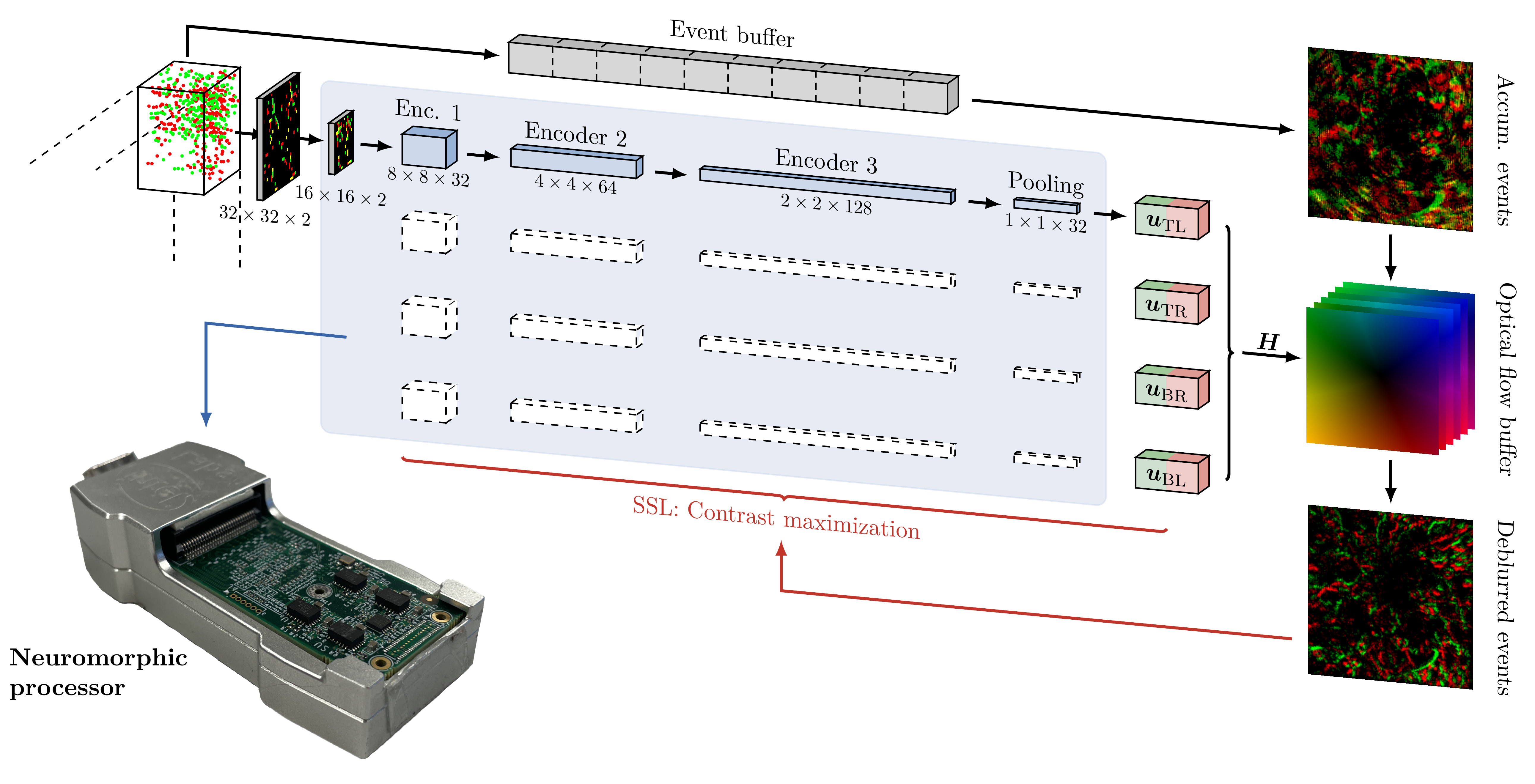}
    \caption{\textbf{Overview of the spiking vision network.} Running at approx. 200~Hz, events are accumulated (max 90 events per corner) and then fed through the vision network consisting of three encoders (kernel size 3x3, stride 2) and a spiking pooling layer. Spikes are decoded into two floats representing flow for that corner. This network is replicated to the three other corners, in order to end up with four corner optical flows vectors. During training, these are used in a homography transformation to derive dense flow, which is then used for the self-supervised loss. The full network is running on the neuromorphic processor during the real-world flight tests.}
    \label{fig:vision}
\end{figure*}

We split the training of our vision-to-control pipeline into two separate frameworks. On the one hand, the vision part of the pipeline, in charge of mapping input events to optical flow, is trained in a self-supervised fashion using the contrast maximization framework \cite{gallego2018unifying,gallego2019focusa}. The idea behind this approach is that, by compensating for the spatiotemporal misalignments among the events triggered by a moving edge (i.e., event deblurring), one can retrieve accurate optical flow information. In this work, we use the formulation proposed in \cite{hagenaars2021selfsupervised} and shown in \figref{fig:vision}. Corner events within non-overlapping temporal windows of 5 milliseconds are processed sequentially by our spiking networks, which provide optical flow estimates at every timestep. Only during training, we use the motion information of the four corners to parameterize a homography transformation that, under the assumption of static planar surface, allows us to retrieve dense optical flow, as in \cite{baker2006parameterizing, detone2016deep, nguyen2018unsupervised, sanket2020evdodgeneta}. Following \cite{hagenaars2021selfsupervised}, we accumulate event and optical flow tuples over multiple timesteps for contrast maximization to be a robust self-supervisory signal, and only compute the deblurring loss function and perform a backward pass through the networks (using backpropagation through time) once 25 milliseconds of event data have been processed. To cope with the non-differentiable spiking function of our neurons, we use surrogate gradients \cite{neftci2019surrogatea}.

\begin{figure*}[!t]
    \centering
    \includegraphics[width=0.9\textwidth]{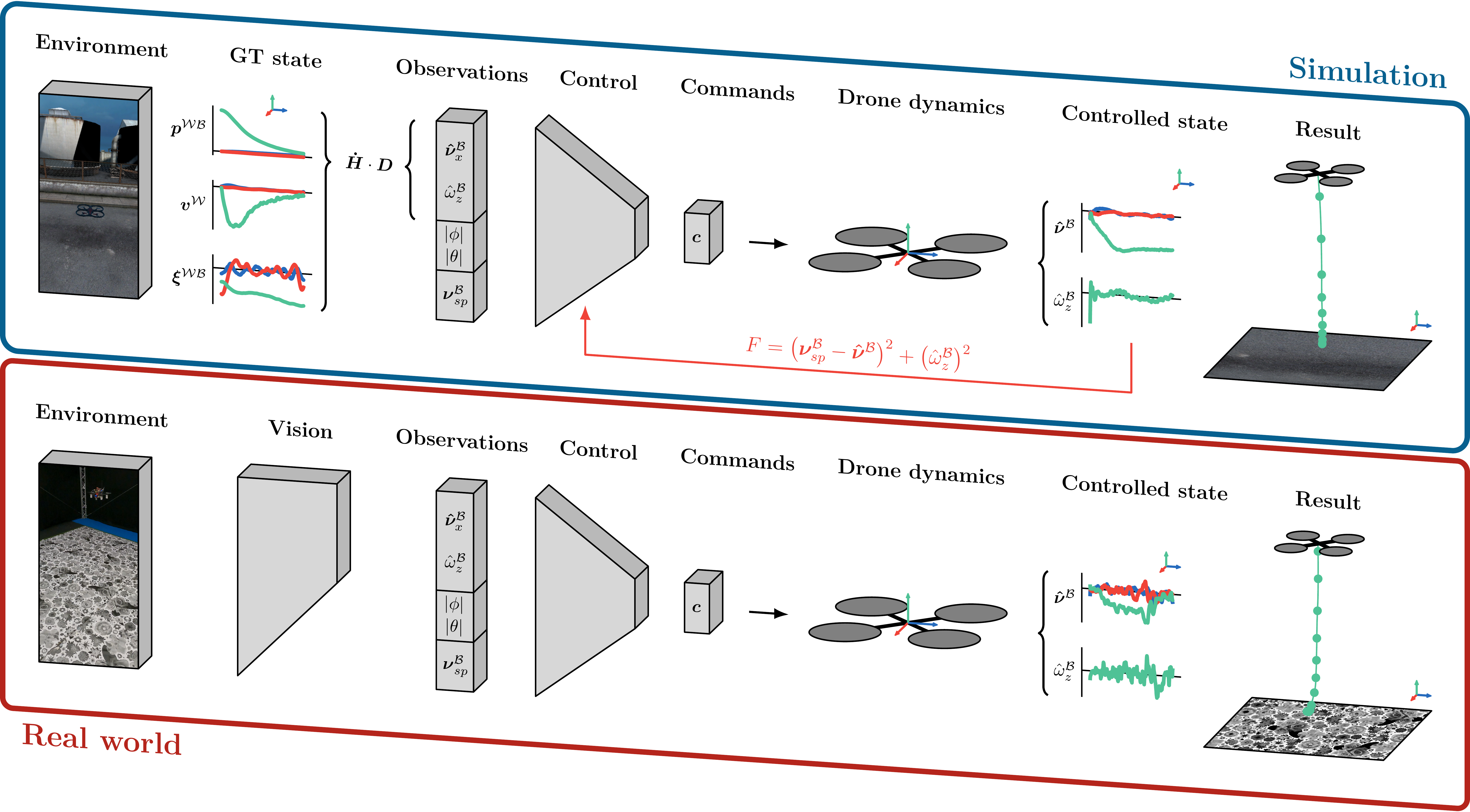}
    \caption{\textbf{Overview of the control pipeline for simulation and real-world tests.} During training in simulation, we construct visual observable observations from ground-truth using the continuous homography transform. The control decoding takes these observables together with roll and pitch and a setpoint to output commands, which control the drone dynamics. We train the controller using evolution based on a fitness signal that quantifies how well the controller can follow setpoints for horizontal and vertical flight. In the real world, we receive corner flows from the vision network, transform these to visual observables and control commands in a single matrix multiplication in order to send low-level control commands thrust and attitude to the autopilot.}
    \label{fig:control}
\end{figure*}

On the other hand, the control part of the network, consisting of a linear mapping from the motion of the four corners to thrust and attitude control commands, is trained in a drone simulator using a genetic algorithm. \figref{fig:control} gives an overview of this. To get around the need to incorporate an event-based vision pipeline in simulation, we use the ground-truth state of the simulated drone to generate the expected corner flows using the continuous homography transform~\cite{ma2004invitation}, and use these to construct the scaled velocity and yaw rate estimates that make up the visual observables of the camera's ego-motion~\cite{longuet-higgins1980interpretation,decroon2013opticflow}. The inputs to the linear control mapping are then these visual observables, absolute roll and pitch (from the drone's inertial measurement unit) and a desired setpoint for the visual observables. The outputs of the controller (i.e., desired collective thrust, pitch and roll angles and yaw rate) are subsequently applied to the drone model in order to control it. During evolution, the fitness of a controller is determined based on the accumulated visual observable error in an evaluation. We evaluate each of the agents in the population on a set of (repeated) setpoints representing horizontal and vertical flight, create offspring through random mutations, and select the best individuals for the next generation. The trained controller is transferred directly to the real robot, without any retraining.

\section*{Results}

Because of the split between the vision and control parts of the pipeline, we can evaluate their performance separately. The estimated corner flows of the vision part are compared against ground truth data obtained from a motion capture system, while the control part is evaluated in simulation. Connecting vision and control together, we then demonstrate the performance of our fully neuromorphic vision-to-control pipeline through real-world flight tests. To further illustrate the robustness of our vision-based state estimation, we perform real-world tests with changing setpoints, and tests in various lighting conditions. Lastly, we compare energy consumption against possible on-board GPU solutions.

\begin{figure*}[!h]
    \centering
    \includegraphics[width=0.985\textwidth]{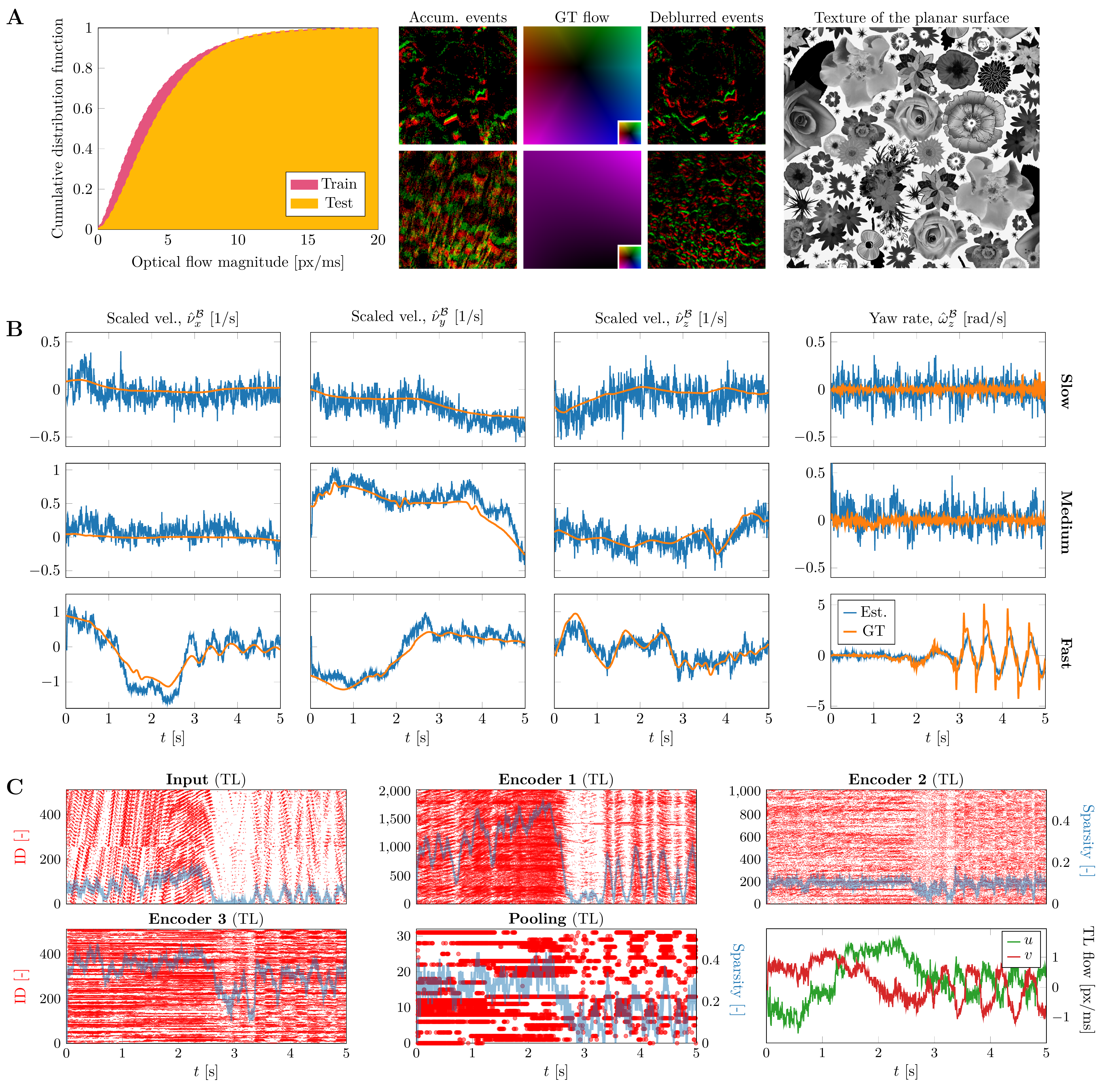}
    \caption{\textbf{Overview of results for the vision-based state estimation.} \textbf{(A)} Qualitative analysis of the training dataset for planar optical flow. Accumulated and blurry events can be deblurred using the ground-truth flow. Events result from a repeating texture on the ground. Flow magnitude is well-distributed across both training and test datasets. \textbf{(B)} Comparison of estimated and ground-truth visual observables for sequences with different motion speeds (slow, medium, fast). \textbf{(C)} Network activity resulting from the top-left-corner events in the fast motion sequence.}
    \label{fig:vision_results}
\end{figure*}

\subsection*{Robust vision-based state estimation}

To prevent reality-gap issues when simulating an event-based camera, we train and evaluate the vision part of our pipeline using real-world event sequences recorded with the same platform (i.e., drone and downward-facing event-based camera) and in the same indoor environment (i.e., static and planar, constant illumination). This dataset consists of approximately 40 minutes of event data, which we split into 25 minutes for training and 15 for evaluation, and its motion statistics are shown in \figref{fig:vision_results}A. In addition to the visual data, the ground truth pose (i.e., position and attitude) of the drone over time is provided at a rate of 180 Hz, and is used solely for evaluation. Examples of this ground truth, which can be converted to dense optical flow using the camera calibration, are shown in \figref{fig:vision_results}A alongside the floor texture of the indoor environment. Note that this dataset is available with the supplementary material of this work.

We train our vision SNN with the self-supervised contrast maximization framework from \cite{hagenaars2021selfsupervised} and a quantization-aware training routine that simulates the neuron and synapse models in the target neuromorphic hardware. Once this is done, we evaluate the performance of our spiking network on the task of planar event-based optical flow estimation using sequences with varying amounts of motion. Qualitative results are presented in \figref{fig:vision_results}B, where the estimated visual observables (constructed from the estimated optical flow vectors at the image corners) are compared to their ground-truth counterparts. These results confirm the validity of our approach. Despite the architectural limitations of the proposed solution (e.g., spike-based processing, limited field of view, only self-recurrency, weight and state quantization) and the fact that it does not have access to ground-truth information during training, it is able to produce optical flow estimates that accurately capture the motion encoded in the input event stream, i.e., the ego-motion of the camera. This is especially remarkable for the shown fast sequence, where towards the end the camera is spinning with approximately 200~deg/s.

In \figref{fig:vision_results}C, we show the internal spiking activity of our vision SNN as it processes the top-left corner of the image space from the fast sequence shown in \figref{fig:vision_results}B, along with the decoded optical flow vectors. These qualitative results provide insight into the type of processing carried out by the proposed architecture, which is spike-based and therefore sparse and asynchronous. Notably, despite the rapid motion in the input sequence, all layers of the SNN maintain activation levels below 50\% of the available neurons. Note that the network was not explicitly trained to promote sparse activations. Furthermore, we can distinguish layers with activity levels that are highly correlated with the input activity (i.e., encoder 1 and pooling), while others rely on their explicit recurrent connections to maintain activity levels that are relatively independent of the input statistics (i.e., encoder 2 and encoder 3). 

In \tabref{tab:AEEchanges}, we provide a quantitative comparison of our solution with other similar recurrent architectures, based on the average endpoint error (EPE) (i.e., Euclidean distance between predicted and ground-truth optical flow vectors). This evaluation not only demonstrates the performance of our spiking network, but also assesses the impact of each mechanism that was incorporated into the pipeline to achieve a solution that could be deployed on Loihi at the target frequency of 200~Hz. Several conclusions can be drawn from these results. Firstly, the ANN outperforms its spiking variants by a large margin, and self-recurrency is the weakest form of explicit recurrency among those tested. Secondly, deploying one architecture to each image corner instead of processing the entire image space at once is beneficial for our architecture, while only having a slight detrimental effect on the baselines. Limiting the number of events that can be processed at once to 90 per corner is also helpful for the evaluated SNNs, as it helps reduce the internal activity levels. Lastly, the incorporation of the Loihi-specific weight and state quantization leads to an error increase for our architecture.

\begin{table*}[!h]
\centering
\caption{\textbf{Quantitative comparison between different architectures}. Bottom right corner, in bold, indicates the eventual architecture. Row-wise architecture choices and column-wise design decisions impact test performance in average endpoint error (EPE).}
\label{tab:AEEchanges}
\resizebox{0.8\textwidth}{!}{%
\begin{tabular}{@{}lccccc@{}}
\toprule
                                 &  \textbf{Full image}   & \textbf{+2x Down.} & \textbf{+Corner crop} & \textbf{+Limit events} & \textbf{+Loihi quant.} \\ \midrule
Conv-GRU ANN                      & +2.45\%    & Best, 0.056 EPE      & +0.94\%       & +2.94\%    & -       \\ \midrule
Conv-RNN SNN                      & +42.48\%    & +41.96\%      & +42.25\%       & +25.48\%    & +25.32\%      \\ \midrule
Self-RNN SNN (ours)                      & +94.16\%    & +76.45\%      & +48.00\%       & +38.74\%    & \textbf{+50.04\%}      \\ 
\bottomrule
\end{tabular}
}
\end{table*}

\begin{figure*}[!h]
    \centering
    \includegraphics[width=0.9\textwidth]{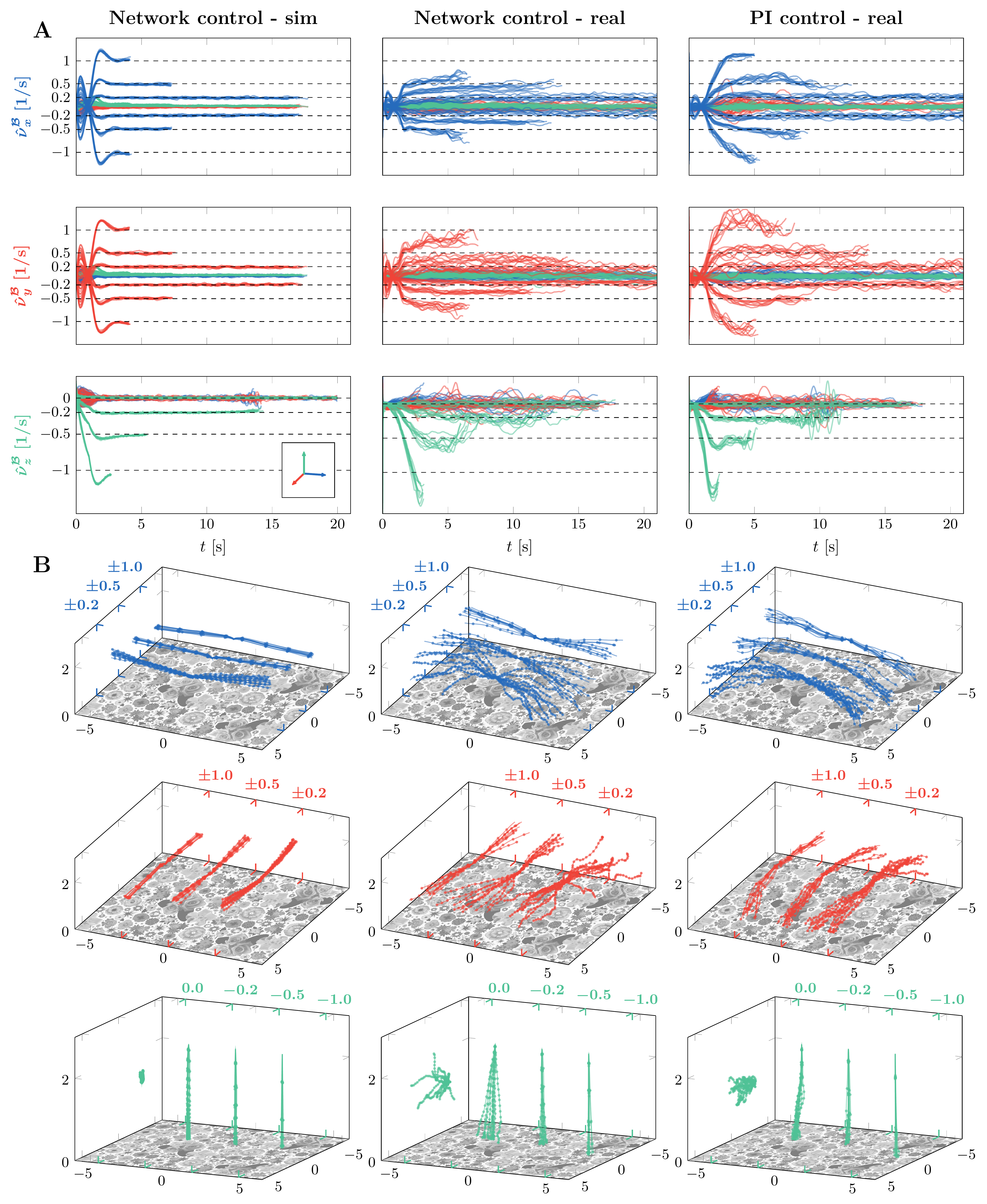}
    \caption{\textbf{Comparison of results obtained in simulation and during real-world flight tests.} \textbf{(A)} Estimated scaled velocities for 16 different setpoints in three axes, across three scenarios: linear network controller in simulation and the real world, and a hand-tuned proportional-integral (PI) controller in the real world. \textbf{(B)} 3D world position trajectories for the same flight tests, offset to accomodate more plots in a single figure. Each cube in \textbf{(B)} matches the plot in the corresponding location in \textbf{(A)}.}
    \label{fig:results1}
\end{figure*}

\subsection*{Control through visual observables: from sim to real}

Separately from the vision part, we train and evaluate the control part of our pipeline. This is a linear mapping from visual observables (i.e., scaled velocities estimate $\bm{\hat{\nu}}^\mathcal{B}$ and yaw rate estimate $\hat{\omega}^\mathcal{B}_z$), absolute roll and pitch and a visual observable setpoint to thrust and attitude commands. A population of these mappings is evolved in simulation for a set of 16 visual observable setpoints. Each scaled velocity setpoint $\bm{\nu}^\mathcal{B}_{sp}$ has at most one nonzero element $\in \{\pm0.2,\pm0.5,\pm1.0\}$~1/s. In other words: they represent hover, vertical flight in the form of landing at three speeds (no ascending flight), and horizontal flight in four directions at three speeds. Unless mentioned otherwise, the setpoint for yaw rate $\smash{\omega^\mathcal{B}_{z,sp} = 0}$. The first column of \figref{fig:results1} shows the performance of the evolved linear network controller in simulation in terms of the estimated scaled velocities $\bm{\hat{\nu}}^\mathcal{B}$ (\figref{fig:results1}A) and the world position $\bm{p}^\mathcal{WB}$ over time (\figref{fig:results1}B) for all setpoints. The controller reaches the setpoint in all cases, and is capable of keeping the scaled velocities for the non-flight direction close to zero. Especially for $\smash{\nu_{*,sp}^\mathcal{B}=\pm1.0}$~1/s, there is overshoot, but this can be expected given that this is a linear mapping without any kind of derivative control.

We get the second column of \figref{fig:results1} by deploying this controller in the real world, and replacing the ground-truth visual observables with those estimated by the vision network. Looking at the scaled velocity plots for the different setpoints, we see that these become less noisy for higher setpoints and faster flight, as can also be seen from the 3D position plots. This is due to the fact that the signal-to-noise ratio of the vision-based state estimation increases with motion magnitude (little motion means most events are due to noise, as can be seen in \figref{fig:vision_results}). Also, the inertia of the drone provides some stability at higher speeds. Overall, the results demonstrate successful deployment of the fully neuromorphic vision-to-control pipeline. Nevertheless, apart from several setpoints (e.g., landings, $\smash{\nu^\mathcal{B}_{\{x,y\},sp} = \pm0.2}$~1/s), the controller is not able to reach the desired setpoint: the steady-state error looks to be proportional to the setpoint magnitude. This can be attributed to the fact that while the controller is a \emph{linear} mapping, the relationship between attitude angle and resulting forward/sideways velocity is \emph{nonlinear} as a result of drag. Providing absolute attitude input to the network, and simulating the drag (as in \cite{dewagter2022sensing}) during training turned out not to be enough to compensate. Furthermore, there can be mismatches between the dynamics of the simulated drone (body characteristics, motor dynamics) with which the controller was trained and the real drone on which the flight tests were performed, even though we abstracted the control outputs to attitude commands. Lastly, inaccuracies of the drag model can also be a source of error (in this case, it seems that drag was higher in reality than in simulation).
 
The third column of \figref{fig:results1} shows the results obtained by connecting a hand-tuned proportional-integral (PI) controller to the vision-based state estimation. We compare this to the linear network controller. Looking at all directions and setpoints, we see that the PI controller reaches the setpoint faster than the network controller. For horizontal flight, the network controller is not at all able to reach the setpoint $\smash{\nu^\mathcal{B}_{\{x,y\},sp} = \pm1.0}$~1/s and only just in the case of $\pm0.5$~1/s, supposedly due to the limitations of linear control. The PI controller does not have this problem, as it can increment its control command to eliminate the steady-state error. For vertical flight, both the network and the PI controller have quite some overshoot for $\nu^\mathcal{B}_{z,sp} = -1.0$~1/s. That this is so obvious, however, has to do with the fact that at such speeds from such heights (i.e., 2.5~m), the drone barely reaches the setpoint before reaching the ground, and therefore has little time to compensate for any overshoot (look at the PI controller for $\nu^\mathcal{B}_{z,sp} = -0.5$~1/s; there overshoot is similar but is corrected shortly after). A slightly lower gain or a derivative term could help here.

\subsection*{Disco, darkness, squares and frisbees: examples of versatility and robustness}

\begin{figure*}[!h]
    \centering
    \includegraphics[width=0.985\textwidth]{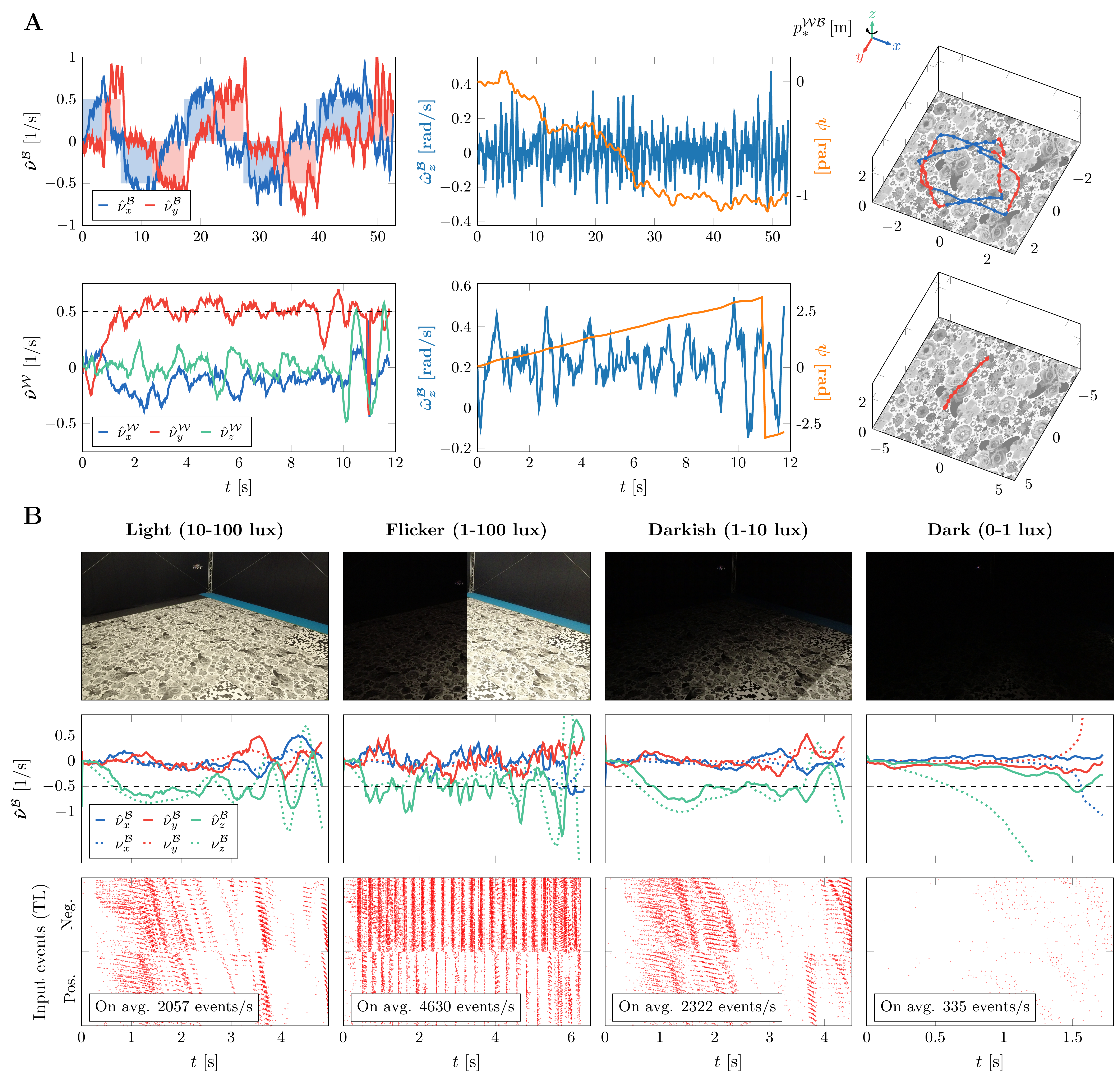}
    \caption{\textbf{Results with vision network and proportional-integral (PI) controller.} \textbf{(A)} Top row: alternating setpoints in X and Y in order to fly a square. Bottom row: rotating the scaled velocity setpoint by the yaw angle leads the drone spinning around its Z-axis while flying in a straight line. \textbf{(B)} Landing experiments with different lighting conditions. While flickering lights lead to many more events, visual observable estimates (and hence control) only diverge when it is so dark that there are almost no events.}
    \label{fig:results3}
\end{figure*}

We can combine the vision-based state estimation with the PI controller to show the versatility and robustness of the former through various other tests, with the benefit of not having to include these in training for the linear network controller. \figref{fig:results3}A shows these tests. The top row displays the user alternating through different scaled velocity setpoints in X and Y (while keeping yaw constant) in order to let the drone fly a square. While the controller is able to reach the desired setpoint quite quickly, allowing for rather sharp corners, there is significant drift in yaw, leading to a slightly rotated second square with respect to the first.

The bottom row of \figref{fig:results3}A shows an experiment in which the drone has to fly in a straight line while spinning around its Z axis like a frisbee. The drone receives a nonzero yaw rate setpoint $\omega^\mathcal{B}_{z,sp} = 0.2$~rad/s. In combination with a setpoint of $\smash{\nu^\mathcal{B}_{y,sp}=0.5}$~1/s this would lead to the drone flying in a circle. To prevent this and achieve the frisbee-like spinning effect, we rotate $\nu^\mathcal{B}_{{y,sp}}$ by the yaw angle. The first and last plot show that this works: despite some drift in X and Z, the setpoints are followed well and the 3D position trajectory is quite straight. The second plot shows that the desired yaw rate is tracked well and that the yaw angle is constantly increasing.

\figref{fig:results3}B shows landing with divergence $\nu^\mathcal{B}_{z,sp}=-0.5$~1/s for various lighting conditions (quantified with lux measurements). The events for the top left corner are shown in the bottom row. The light and darkish settings look alike, but flickering lights lead to a large increase in events, while the darkest setting gives almost no events. As the middle row of plots shows, despite the challenging light conditions, the controller is able to track the setpoint (black dashed line) quite well, and the estimated scaled velocities approximate their ground thruths. Only the darkest setting poses a real problem for the state estimation: in that case, the estimated scaled velocities $\bm{\hat{\nu}}^\mathcal{B}$ diverge too much from the ground truth scaled velocities $\bm{{\nu}}^\mathcal{B}$ to perform a successful landing. 

\subsection*{Improved inference speed and energy consumption on neuromorphic hardware}

\begin{table*}[!h]
\centering
\caption{\textbf{Approximate energy and power characteristics for various devices on three sequences: slow, medium and fast.} On average, slow has 28.6 events/inf, medium has 106.9 events/inf, and fast has 186.6 events/inf. Delta power is the difference between idle and running (total) power, and is used to compute energy per inference. Dynamic power is the power needed for switching and short-circuiting, while static power is due to leakage; together they sum to running (total) power as well. Nahuku is a board with 32 Loihi chips (Kapoho Bay has 2). A Nahuku configuration where no spikes are sent and only chips and cores are allocated (no synapses) is included as `empty'. Jetson Nano has a low-power (5W) and high-power (10W) mode.}
\label{tab:energy}
\resizebox{0.9\textwidth}{!}{%
\begin{tabular}{@{}lrccccccc@{}}
\toprule
\textbf{Device}                                 &  \textbf{Seq.}   & \textbf{Static [W]} & \textbf{Dynamic [W]} & \textbf{Idle [W]} & \textbf{Running [W]} & \textbf{Delta [W]} & \textbf{Inf/s} & \textbf{\textmu J/Inf} \\ \midrule
Nahuku 32 (empty)                      & any    & 0.861      & 0.040       & 0.897    & 0.901       & 0.004     & 60496 & 0.071      \\ \midrule
\multirow{3}{*}{Nahuku 32} & slow   & 0.899      & 0.048       & 0.935    & 0.947       & 0.012     & 1637  & 7.165      \\
                                       & medium & 0.900      & 0.044       & 0.936    & 0.945       & 0.008     & 411   & 20.602     \\
                                       & fast   & 0.902      & 0.043       & 0.938    & 0.946       & 0.007     & 274   & 27.207     \\ \midrule
\multirow{3}{*}{Jetson Nano (5W)}      & slow   & -          & -           & 1.053  & 2.229       & 1.177     & 14     & 86111.152 \\
                                       & medium & -          & -           & 1.027   & 2.245       & 1.218     & 14     & 85576.977 \\
                                       & fast   & -          & -           & 1.030    & 2.238       & 1.208     & 14     & 86187.999 \\ \midrule
\multirow{3}{*}{Jetson Nano (10W)}     & slow   & -          & -           &  1.043   & 2.974       & 1.931     & 26     & 75246.618 \\
                                       & medium & -          & -           &  1.058   & 2.981       & 1.923     & 26     & 75347.505 \\
                                       & fast   & -          & -           &  1.044   & 2.991       & 1.947     & 25     & 76522.389 \\ \bottomrule
\end{tabular}
}
\end{table*}

\tabref{tab:energy} shows a comparison in terms of power/energy and runtime between the Loihi neuromorphic processor and an NVIDIA Jetson Nano for running the vision network on sequences with varying amounts of motion and hence varying input event density. The SNN runs in hardware on Loihi and in software (PyTorch) on Jetson Nano. The tests for Loihi were performed on a Nahuku board, which contains 32 Loihi chips. We confirmed, insofar possible, that using two chips on Nahuku is representative of a Kapoho Bay (at least in terms of execution time), which is the two-chip form factor used on the drone. Still, neither of these benchmarks is completely representative of the tests performed in the real world: the benchmarks use data already loaded in memory, and therefore only quantify the processing by the network without any bottlenecks or impacts due to I/O and preprocessing, whereas the flight tests involve streaming event data that is coming in and is being processed in an online fashion. This shows in Loihi's execution frequencies in \tabref{tab:energy}, which are well above the 200 Inf/s achieved during flight tests.

Because Jetson Nano does not provide static and dynamic power components, we compare the difference between idle and running power, and use that to compute energy per inference. Loihi, depending on the sequence, outperforms Jetson Nano by three to four orders of magnitude, providing a one to two orders of magnitude improvement in execution frequency. Furthermore, the benefits of neuromorphic processing show in Loihi's increasing execution frequency as event sparsity increases (from fast to slow motion sequences). Note that a GPU like Jetson Nano is not optimized to simulate SNNs, and a more efficient implementation could be obtained by running a feedforward ANN with multiple temporal windows of events as input.

\section*{Discussion and conclusion}

We presented the first fully neuromorphic vision-to-control pipeline for controlling a freely flying drone. Specifically, we trained a spiking neural network that takes in high-dimensional raw event-based camera data and produces low-level control commands. Real-world experiments demonstrated a successful sim-to-real transfer: the drone can accurately follow various ego-motion setpoints, performing hovering, landing, and lateral maneuvers\textemdash even under constant yaw rate. 

Our study confirms the potential of a fully neuromorphic vision-to-control pipeline by running on board with an execution frequency of 200~Hz, spending only 27~\textmu J per network inference. However, there are still important hurdles on the way to reaping the full system benefits of such a pipeline, embedding it on extremely lightweight (e.g., <30~g) drones. 

For reaching the full potential, the entire drone sensing, processing, and actuation hardware should be neuromorphic, from its accelerometer sensors to the processor and motors. Such hardware is currently not available, so we have limited ourselves to the vision-to-control pipeline, ending at thrust and attitude commands. Concerning the neuromorphic processor, the biggest advancement could come from improved I/O bandwidth and interfacing options. The current processor could not be connected to the event-based camera directly via AER, and with our advanced use case, we reached the limits of the number of spikes that can be sent to and received from the neuromorphic processor at the desired high execution frequency. This is also the reason that we have limited ourselves to a linear network controller: the increase in input spikes needed to encode the setpoint and attitude inputs would substantially reduce the execution frequency of the pipeline. Ultimately, further gains in terms of efficiency could be obtained when moving from digital neuromorphic processors to analog hardware, but this will pose even larger development and deployment challenges. 

Despite the above-mentioned limitations, the current work presents a substantial step towards neuromorphic sensing and processing for drones. The results are encouraging, because they show that neuromorphic sensing and processing may bring deep neural networks within reach of small autonomous robots. In time this may allow them to approach the agility, versatility and robustness of animals such as flying insects.

\section*{Materials and methods}

Here, we explain the main components of the proposed fully-neuromorphic vision-to-control pipeline, starting with the neuron model of our SNN and how this is trained in a self-supervised fashion using real event camera data. Next, we describe how the vision-based state estimate can be used for navigation, and how we train a controller on top of it. Finally, we discuss the real-world tests and hardware and the performed energy benchmarks.

\subsection*{Neuromorphic state estimation: spiking, sequential processing}

In this study, we utilize a spiking neuron model based on the current-based leaky-integrate-and-fire (CUBA-LIF) neuron, whose membrane potential $U$ and synaptic input current $I$ at timestep $t$ can be written as:
\begin{align}
    U^{t}_{i} &= \tau_U (1 - S^{t-1}_{i}) U^{t-1}_{i} + I^{t}_{i} \\
    I^{t}_{i} &= \tau_I I^{t-1}_{i} + \sum_j w_{ij}^{\text{ff}} S_j^{t} + w_{ii}^{\text{rec}} S_i^{t-1}
\end{align}
where $j$ and $i$ denote presynaptic (input) and postsynaptic (output) neurons within a layer, $S \in \{0, 1\}$ a neuron spike, and $w^{\text{ff}}$ and $w^{\text{rec}}$ feedforward and self-recurrent connections (if any), respectively. The decays (or leaks) of the two internal state variables of this neuron model are learned, and are denoted by $\tau_U$ and $\tau_I$. A neuron fires an output spike if the membrane potential exceeds a threshold $\theta$, which is also learned. The firing of a spike triggers a hard reset of the membrane potential. Note that, in this work, all neurons within a layer share the same decays and firing threshold.

Neurons on the Loihi neuromorphic processor also follow the CUBA-LIF model \cite{davies2018loihi}, however, several considerations must be taken into account to accurately simulate these on-chip neurons. Firstly, the two states variables are quantized in the integer domain. Hence, the parameters associated with these variables are also quantized in the same way: $w\in[-256\ ..\ 256 - \Delta w]$ with $\Delta w$ being the quantization step for the synaptic weights, $\tau_{\{U,I\}}\in[0\ ..\ 4096]$ for the decays, and $\theta\in[0\ ..\ 131071]$ for the threshold. We follow this quantization scheme with $\Delta w=8$ (6-bit weights) in the simulation and training of our neural networks. Secondly, to emulate the arithmetic left (bit) shift operations carried out by the processor when updating the neuron states, we perform a rounding towards zero operation after the application of the decays. Taking these aspects into consideration, we obtain a matching score of 100\% between the simulated and the on-chip spiking neurons. We use quantization-aware training (quantized forward pass, floating-point backward pass) to minimize the performance loss of our SNN when deployed on Loihi.

As surrogate gradient for the spiking function $\sigma$, we opt for the derivative of the inverse tangent $\smash{\sigma'(x) = \text{aTan}' = 1/(1 + \gamma x^2)}$ \cite{fang2021incorporating}, with $\gamma=10$ being the surrogate width and $\smash{x = u - \theta}$.

\subsection*{Neuromorphic state estimation: planar homography}

Assuming that $\boldsymbol{x}=\left[x, y, 1\right]^T$ and $\boldsymbol{x}'=\left[x', y', 1\right]^T$ are two undistorted corresponding points from a planar scene expressed in homogeneous coordinates and captured by a pinhole camera at different time instances, a planar homography transformation is a linear projective transformation that maps $\boldsymbol{x}\leftrightarrow\boldsymbol{x}'$ such that:
\begin{align}\label{eq:homo}
    \lambda 
    \begin{bmatrix}
    x'\\
    y'\\
    1
    \end{bmatrix}
    =
    \boldsymbol{H}
    \begin{bmatrix}
    x\\
    y\\
    1
    \end{bmatrix};\quad \text{with } \boldsymbol{H}=
    \begin{bmatrix}
    h_{11} & h_{12} & h_{13}\\
    h_{21} & h_{22} & h_{23}\\
    h_{31} & h_{32} & 1
    \end{bmatrix}
\end{align}
where $\boldsymbol{H}$ is a 3x3 non-singular matrix, further referred to as the homography matrix, which is characterized by eight degrees of freedom and is defined up to a scale factor $\lambda$. 

From \eqnref{eq:homo}, we can formulate $\boldsymbol{A}_k \boldsymbol{h} = \boldsymbol{b}_k$, an underdetermined system of linear equations for the $k$-th point correspondence, where:
\begin{align}
    \boldsymbol{A}_k &=  
    \begin{bmatrix}
    x & y & 1 & 0 & 0 & 0 & -x'x & -x'y \\
    0 & 0 & 0 & x & y & 1 & -y'x & -y'y\\
    \end{bmatrix}\\
    \boldsymbol{h} &= 
    \begin{bmatrix}
    h_{11} & h_{12} & h_{13} & h_{21} & h_{22} & h_{23} & h_{31} & h_{32}
    \end{bmatrix}^T\\
    \boldsymbol{b}_k &=
    \begin{bmatrix}
    x' & y'
    \end{bmatrix}^T
\end{align}
As shown in \figref{fig:vision}A, our vision network predicts the displacement of the corner pixels in a certain time window. Using this information, we can solve for the components of the homography matrix through $\boldsymbol{h}=\boldsymbol{A}^{-1}\boldsymbol{b}$, with $\boldsymbol{A}$ and $\boldsymbol{b}$ being the result of the concatenation of the individual $\boldsymbol{A}_k$ and $\boldsymbol{b}_k$ of each point correspondence $\forall k\in\{\mathrm{TL},\mathrm{TR},\mathrm{BR},\mathrm{BL}\}$. This approach is referred to as the four-point parametrization of the homography transformation \cite{baker2006parameterizing}, and it has proved to be successful in the event-camera literature for robotics applications \cite{sanket2020evdodgeneta, ozawa2022accuracya}. However, to the best of our knowledge, it has never been formulated for SNNs trained with self-supervised learning.

Once the homography matrix is estimated, we can estimate a dense (i.e., per-pixel) optical flow map as follows:
\begin{align}\label{eq:flow}
    \boldsymbol{u}(\boldsymbol{x}, \boldsymbol{H})=
    \begin{bmatrix}
    u(\boldsymbol{x}, \boldsymbol{H})\\
    v(\boldsymbol{x}, \boldsymbol{H})
    \end{bmatrix}
    =
    \boldsymbol{H}
    \begin{bmatrix}
    x\\
    y
    \end{bmatrix} - 
    \begin{bmatrix}
    x\\
    y
    \end{bmatrix}
\end{align}
which encodes the displacement of pixel $\boldsymbol{x}$ in the time window of $\boldsymbol{H}$.

\subsection*{Neuromorphic state estimation: self-supervised learning}

To train our spiking architecture to estimate the displacement of the four corner pixels in a self-supervised fashion, we use the contrast maximization framework for motion compensation \cite{gallego2018unifying, gallego2019focusa}. Assuming constant illumination, accurate optical flow information is encoded in the spatiotemporal misalignments among the events triggered by a moving edge (i.e., blur). To retrieve it, one has to learn to compensate for this motion (i.e., deblur the event partition) by transporting the events through space and time. Once we get a per-pixel optical flow estimate $\boldsymbol{u}(\boldsymbol{x}, \boldsymbol{H})$ from \eqnref{eq:flow}, we can propagate the events to a reference time $t_{\text{ref}}$ through the following linear motion model:
\begin{equation}\label{eqn:motionmodel}
    \boldsymbol{x}'_i=\boldsymbol{x}_i + (t_{\text{ref}} - t_i)\boldsymbol{
    u}(\boldsymbol{x}_i, \boldsymbol{H})
\end{equation}
and the result of aggregating the propagated events is referred to as the image of warped events (IWE) at $t_{\text{ref}}$.

As loss function, we use the reformulation from \cite{hagenaars2021selfsupervised} of the focus objective function based on the per-pixel and per-polarity average timestamp of the IWE \cite{mitrokhin2018eventbased, zhu2019unsuperviseda}. The lower this metric, the better the event deblurring and hence the more accurate the estimated optical flow. We generate an image of the per-pixel average timestamp for each polarity $p'$ via bilinear interpolation:
\begin{align}\label{eqn:timeimage}
\begin{aligned}
T_{p'}(\boldsymbol{x}{;}\boldsymbol{u} |t_{\text{ref}}) &= \frac{\sum_{j} \kappa(x - x'_{j})\kappa(y - y'_{j})t_{j}}{\sum_{j} \kappa(x - x'_{j})\kappa(y - y'_{j})+\epsilon}\\\kappa(a) &= \max(0, 1-|a|)\\
j = \{i \mid p_{i}=&\ p'\}, \hspace{15pt}p'\in\{+,-\}, \hspace{15pt} \epsilon\approx 0
\end{aligned}
\end{align}

Following \cite{hagenaars2021selfsupervised}, we first scale the sum of the squared temporal images resulting from the warping process with the number of pixels with at least one warped event:
\begin{equation}\label{eq:scaling}
\mathcal{L}_{\text{contrast}}(t_{\text{ref}}) = \frac{\sum_{\boldsymbol{x}} T_{+}(\boldsymbol{x}{;}\boldsymbol{u} |t_{\text{ref}})^2 + T_{-}(\boldsymbol{x}{;}\boldsymbol{u} |t_{\text{ref}})^2}{\sum_{\boldsymbol{x}}\left[n(\boldsymbol{x}') > 0\right] + \epsilon}
\end{equation}
where $n(\boldsymbol{x}')$ denotes a per-pixel event count of the IWE.

As in \cite{zhu2019unsuperviseda, paredes-valles2021back, hagenaars2021selfsupervised}, we perform the warping process both in a forward ($t_{\text{ref}}^{\text{fw}}$) and in a backward fashion ($t_{\text{ref}}^{\text{bw}}$) to prevent temporal scaling issues during backpropagation. The total loss used to train our event-based optical flow networks is then given by:
\begin{align}\label{eqn:flow_loss}
\mathcal{L}_{\text{contrast}} &= \mathcal{L}_{\text{contrast}}(t_{\text{ref}}^{\text{fw}}) + \mathcal{L}_{\text{contrast}}(t_{\text{ref}}^{\text{bw}})\\
\mathcal{L}_{\text{flow}} &= \mathcal{L}_{\text{contrast}} + \lambda_{\mathcal{L}} \mathcal{L}_{\text{smooth}}\label{eqn:final_loss}
\end{align}
where $\mathcal{L}_{\text{smooth}}$ is a Charbonnier smoothness prior \cite{charbonnier1994two} applied in the temporal domain to subsequent per-corner optical flow estimates, while $\lambda_{\mathcal{L}}$ is a scalar balancing the effect of the two losses. We empirically set this weight to $\lambda_{\mathcal{L}}=0.1$.

As discussed in \cite{hagenaars2021selfsupervised}, there has to be enough linear blur in the input event partition for this loss function to be a robust supervisory signal \cite{gallego2019focusa, stoffregen2019eventa}. Since we process the event stream sequentially, with only a few events being considered at each forward pass, we define the so-called training partition $\smash{\boldsymbol{\varepsilon}^{\text{train}}_{k\rightarrow k+K}\doteq\{(\boldsymbol{\varepsilon}^{\text{inp}}_{i}, \hat{\boldsymbol{u}}_i)\}_{i=k}^{K}}$, which is a buffer that gets populated every forward pass with the input events and their corresponding optical flow estimates. This is illustrated in \figref{fig:vision}A. At training time, we perform a backward pass with the content of the buffer using backpropagation through time once it contains 5 successive event-flow tuples (i.e., 25 milliseconds of event data), after which we update the model parameters, detach its states from the computational graph, and clear the buffer. We use a batch size of 16 and train until convergence with the Adam optimizer \cite{kingma2017adam} and a learning rate of $1\mathrm{e}{}$-$4$.

\subsection*{From a vision-based state estimate to control}

The corner flows $\left[\bm{u}_\mathrm{TL},\bm{u}_\mathrm{TR},\bm{u}_\mathrm{BR},\bm{u}_\mathrm{BL}\right]^T \in \mathbb{R}^{8\times 1}$ resulting from the vision-based state estimation can be used to control the drone. More specifically, we can transform the corner flows to visual observable estimates~\cite{decroon2013opticflow}, consisting of scaled velocities $\bm{\hat{\nu}}^\mathcal{C} \in \mathbb{R}^{3\times 1}$ and yaw rate $\hat{\omega}^\mathcal{C}_z$ in the camera frame $\mathcal{C}$, as follows~\cite{longuet-higgins1980interpretation}:
\begin{align}
    \bm{u}_k &= -
    \begin{bmatrix}
        \nu_x^\mathcal{C}\\
        \nu_y^\mathcal{C}
    \end{bmatrix}
    + \nu^\mathcal{C}_z\bm{x}_k +
    \begin{bmatrix}
        \omega_z^\mathcal{C}\\
        -\omega_z^\mathcal{C}
    \end{bmatrix}
    \circ \bm{x}_k \label{eq:visobs}
\end{align}
where $\circ$ denotes the Hadamard (element-wise) product and $\bm{x}_k = [x,y]^T$ is the projection of the world points in the corners of the field of view onto the pixel array.
Furthermore, it is assumed that 1) the scene is static and planar, 2) angles in pitch and roll are small and 3) optical flow is derotated in pitch and roll. Inverting this relation for all four corners allows us to do a least-squares estimation of the scaled velocities $\bm{\hat{\nu}}^\mathcal{C}$ and the yaw rate $\hat{\omega}^\mathcal{C}_z$, which can then be transformed to the body frame $\mathcal{B}$. To perform control, we can let a user select setpoints $\bm{\nu}_{sp}^\mathcal{B}$ and $\omega_{z,sp}^\mathcal{B}$, and use a trained or manually tuned controller to minimize the difference between the estimated visual observables and their setpoints.

\begin{figure}[!h]
    \centering
    \includegraphics[width=\linewidth]{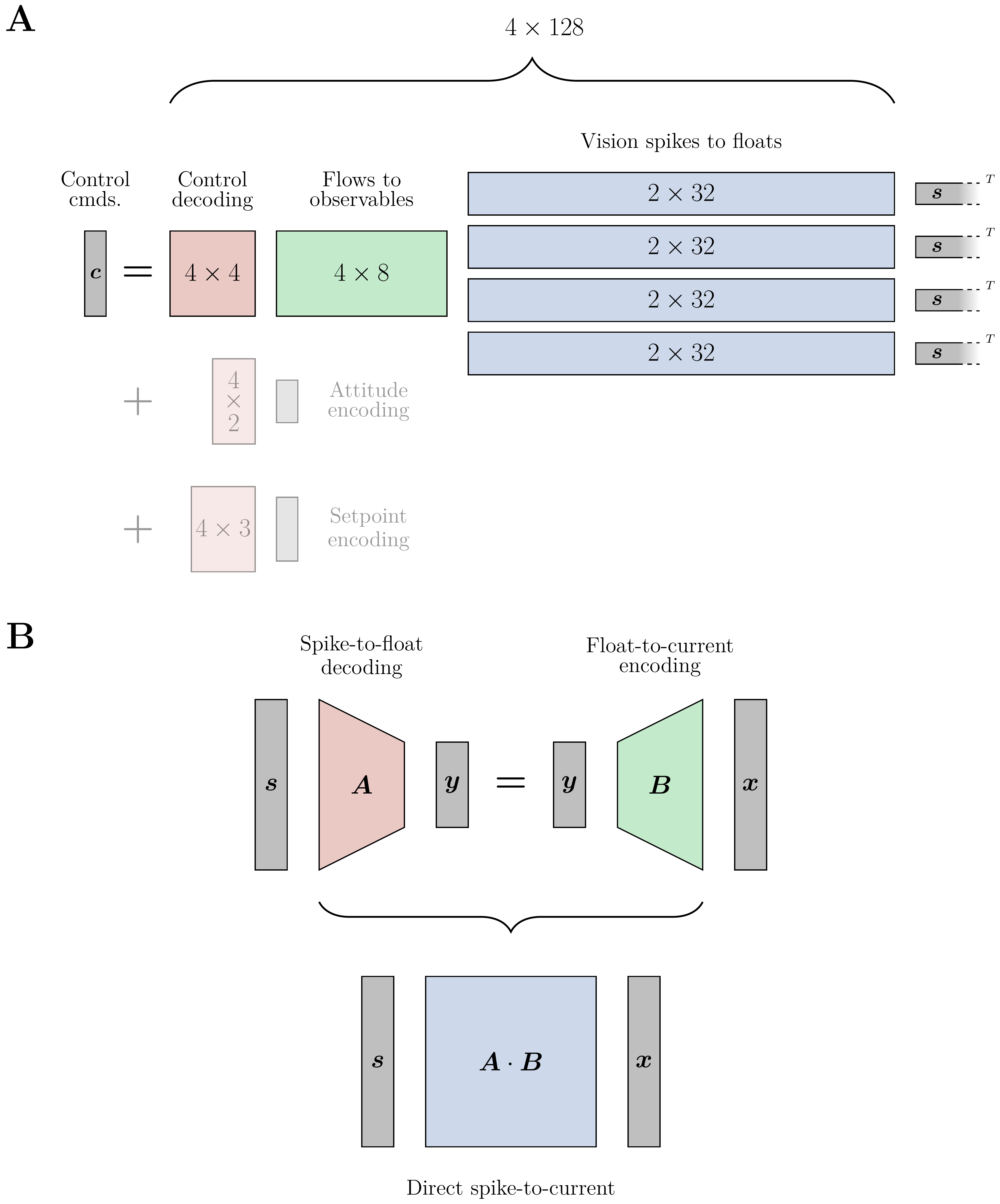}
    \caption{\textbf{Merging linear transformations.} \textbf{(A)} We go directly from output spikes $\bm{s}$ of the vision network to control commands $\bm{c}$ in a single linear decoding by multiplying the involved linear transformation matrices. \textbf{(B)} The same principle can be applied to connect two separately trained spiking networks in a spiking manner, from spikes $\bm{s}$ to currents $\bm{c}$, suitable for neuromorphic hardware.}
    \label{fig:merge}
\end{figure}

Because \eqnref{eq:visobs} is a linear transformation, it can be `merged' with other transformations if these are also linear. This holds for the decoding from spikes to corner flows in the vision SNN, meaning that we can use a single linear transformation from spikes to control commands if we use a linear controller. In a similar fashion, we can use this idea to connect separately trained SNNs, merging their linear decodings and encodings. If both are implemented on neuromorphic hardware, this would mean that no off-chip transfer is necessary. \figref{fig:merge} illustrates these concepts.

\subsection*{Training control in simulation}

We perform control by linearly transforming the visual observable estimates $\smash{\bm{\hat\nu}^\mathcal{B} \in \mathbb{R}^{3\times 1}}$ and $\smash{\hat\omega_z^\mathcal{B}}$, the drone's absolute roll $\lvert\phi\rvert$ and pitch $\lvert\theta\rvert$ and the scaled velocity setpoint $\smash{\bm{\nu}^\mathcal{B}_{sp} \in \mathbb{R}^{3\times1}}$ to a control command $\smash{\bm{c} \in \mathbb{R}^{4\times 1}}$, which consists of an upward, mass-normalized collective thrust offset from hover $\bar{f}_{0,c}$ in the body frame $\mathcal{B}$, a roll angle $\phi_c$ and pitch angle $\theta_c$, and a yaw rate $\smash{\omega^\mathcal{B}_{z,c}}$, in order to reach a certain setpoint of scaled velocities $\smash{\bm{\nu}_{sp}^\mathcal{B}}$ and yaw rate $\smash{\omega_{z,sp}^\mathcal{B}}$ (always 0).

The control part is trained separately from the vision part because of the cost of accurately simulating event-based camera inputs (this needs subpixel displacements between frames, hence high frame rate for fast motion). Simulation is done with a modified version of the drone simulator Flightmare \cite{song2020flightmare}. To mimic the output of the vision-based state estimation network, we first compute the ground-truth continuous homography \cite{ma2004invitation,zhong2020direct} from the state of the drone:
\begin{align}
    \bm{\dot{H}} &= \bm{K} \left( [\bm{\omega}^\mathcal{C}]_\times + \frac{1}{p^\mathcal{WC}_z}\bm{v}^\mathcal{C}(\bm{e}^\mathcal{W}_{-z})^T \right) \bm{K}^{-1}
\end{align}
where $\bm{\dot{H}}$ is the continuous homography, $\bm{K}$ is the camera intrinsic matrix, $[\bm{\omega}^\mathcal{C}]_\times \in \mathbb{R}^{3\times 3}$ is a skew-symmetric matrix representing infinitesimal rotations, $p_z^\mathcal{WC}$ is the Z-component of the position vector from the world frame $\mathcal{W}$ to the camera frame $\mathcal{C}$ (representing perpendicular distance from the ground plane to the camera), $\bm{v}^\mathcal{C}$ is the velocity of the camera, and $\bm{e}_{-z}^\mathcal{W}$ is the unit vector in the negative Z-direction of the world frame. To obtain angular rates and velocities in the camera frame, we use the camera extrinsics, consisting of a rotation $\bm{R}^\mathcal{CB}$ and a translation $\bm{T}^\mathcal{CB}$:
\begin{align}
    \bm{\omega}^\mathcal{C} &= \bm{R}^\mathcal{CB} \bm{\omega}^\mathcal{B} \\
    \bm{v}^\mathcal{C} &= \bm{R}^\mathcal{CB} \left( \bm{v}^\mathcal{B} + [\bm{\omega}^\mathcal{B}]_\times \bm{T}^\mathcal{CB} \right)
\end{align}

Next, we use the continuous homography to get the flow of the four corners~\cite{ma2004invitation,zhong2020direct}:
\begin{align}
    \bm{u}_k &= -\left(\bm{1} - \bm{x}_k(\bm{e}^\mathcal{W}_{-z})^T\right)\bm{\dot{H}}\bm{x}_k
\end{align}
where $\bm{1}$ is the identity matrix, and $\bm{x_k} = [x,y,1]^T$ is the projection of the world points in the corners of the field of view onto the pixel array in homogeneous coordinates (so, $\bm{x}_{BL} = [0,180,1]^T$ and $\bm{x}_{TR}=[180,0,1]^T$, note the difference with respect to \eqnref{eq:visobs}). We add $\mathcal{N}(0, 0.025)$ noise to the flows $\bm{u}_k$ (based on a characterization of the vision SNN). \eqnref{eq:visobs} is subsequently used to go from corner flows to visual observables in the camera frame, which is then transformed back to the body frame for control.

We use a mutation-only genetic algorithm with a population size of 100 to evolve the weights of the linear controller matrix $\in \mathbb{R}^{4\times 9}$, initialized as $\mathcal{U}(-0.1, 0.1)$. More specifically, we generate offspring by adding mutations drawn from $\mathcal{N}(0, 0.001)$ to all parameters of each parent and then evaluate the fitness of both parents and offspring. The next generation is comprised of the best 100 individuals, and we repeat this process until convergence (approx. 25k generations). We use Flightmare to assess fitness at flying various visual observable setpoints: every individual is evaluated across a set of 16 setpoints, with each scaled velocity setpoint $\bm{\nu}^\mathcal{B}_{sp}$ having at most one nonzero element $\in \{\pm0.2,\pm0.5,\pm1.0\}$~1/s, skipping the positive setpoints for the Z-direction, and including hover. The yaw rate setpoint is set to $\omega_{z,sp}^\mathcal{B}=0$ for all. Each setpoint is repeated ten times, meaning a total of 160 evaluations per individual. Fitness $F$ is computed as:
\begin{align}
    F &= \frac{1}{N_\mathrm{eval}} \sum_{i \in N_\mathrm{eval}} \sum_{j \in N_{\mathrm{steps}}} \bm{w} \cdot \left( \bm{\nu}^\mathcal{B}_{sp,i} -
    \begin{bmatrix}
        \hat{\nu}_x^\mathcal{B}\\
        \hat{\nu}_y^\mathcal{B}\\
        \nu_z^\mathcal{W}
    \end{bmatrix}_j\,
    \right)^2 + \left(\hat{\omega}_z^\mathcal{B}\right)^2
\end{align}
Here, $N_{\mathrm{eval}}=160$ is the number of evaluations, $N_\mathrm{steps}=1000$ is the number of steps per episode, and $\bm{w}=[1, 1, w_z]^T$ is a vector weighing the fitness for different axes, where we set $w_z=10$ for setpoints where $\nu^\mathcal{B}_{z,sp}=0$. Note that, for the Z-direction, we use the ground-truth scaled velocity in the world frame $\nu_z^\mathcal{W}$ instead of the one in the body frame, as the latter is zero in the case of the drone ascending or descending at a slope equal to its attitude, and would hence go unpunished, leading to extra vertical drift. Furthermore, if the agent goes out of bounds or crashes before the end of the episode, it will be reset without any additional fitness penalty.

We use domain randomization~\cite{tobin2017domain} to obtain a more robust controller and reduce the reality gap: for each of the ten repeats, a random constant bias $\mathcal{U}(-0.001, 0.001)$~rad is added to the absolute pitch and roll received by the control layer. This bias is shared among the population to keep things fair. Furthermore, for each of the 160 evaluations per individual, we randomly vary the initial position $\bm{p}^\mathcal{WB} = [0,0,2]^T + \mathcal{U}(-1, 1)$ m (except for horizontal setpoints, where we fix $p^\mathcal{WB}_{z}$ to 1.5~m, due to the linear nature of the controller we have here, as explained later), initial velocity $\bm{v}^\mathcal{WB} \in \mathcal{U}(-0.02, 0.02)$~m/s, initial attitude quaternion $\bm{q}^\mathcal{WB} \in \mathcal{U}(-0.02, 0.02)$ (normalized), and initial angular rates $\bm{\omega}^\mathcal{B} \in \mathcal{U}(-0.02, 0.02)$~rad/s.

We modify Flightmare to include drag $\bm{f}_\mathrm{drag}$ occurring as a result of translational motion, and we take it to be acting in the so-called `flat-body` frame $\mathcal{B'}$, which is the body frame rotated by the roll and pitch of the drone, such that the Z-axis is aligned with the world Z-axis. Following~\cite{dewagter2022sensing}, we use a drag model that is linear with respect to velocity in X and Y, but using a drag coefficient $k_{v,x}=k_{v,y}=0.5$. This results in the following:
\begin{align}
    \bm{f}_\mathrm{drag}^\mathcal{B} &= -\bm{R}^\mathcal{BB'}
    \left(
    \begin{bmatrix}
        k_{v,x} \\
        k_{v,y} \\
        0
    \end{bmatrix}
    \circ \bm{R}^\mathcal{B'B}\bm{v}^\mathcal{B}
    \right)
\end{align}

The outputs of the linear controller $\bm{c}\in \mathbb{R}^{4\times1}$ are clamped to $[-1,1]$ and fed to different parts of the cascaded low-level (thrust, attitude and rate) controllers. To accommodate some of the shortcomings of the linear controller, we compensate thrust for the attitude of the drone.

All dynamics equations are integrated with 4$^{\mathrm{th}}$-order Runge-Kutta with a timestep of 2.5~ms. The frequency of the simulation is 50~Hz. All remaining details can be found in the supplementary materials.

\subsection*{From simulation to the real world}

We achieve successful sim-to-real transfer through several strategies. One is domain randomization~\cite{tobin2017domain}, which we do by adding noise to the observed flows, through a random bias on the attitude estimate, and by varying the initial conditions of the simulated quadrotor. Another is abstraction~\cite{scheper2017abstraction}, which we do by making use of low-level controllers to go from thrust and attitude to rotor speeds and by calibrating the attitude and hover thrust biases before each flight to make sure they are small/zero (as in the simulator). Finally, we smooth (low-pass filter) and scale the computed visual observables and tune the gains scaling the control layer outputs in the real world.

\eqnref{eq:visobs} is used to transform the corner flows coming from the vision network into visual observables (scaled velocities $\bm{\hat{\nu}}^\mathcal{B}$ and yaw rate $\hat{\omega}^\mathcal{B}_z$), absolute roll $\lvert\phi\rvert$ and pitch $\lvert\theta\rvert$ are taken from the drone's accelerometers, and the scaled velocity setpoint $\bm{\nu}^\mathcal{B}_{sp}$ is provided by the user. The yaw rate setpoint $\smash{\omega_{z,sp}^\mathcal{B}=0}$ is fixed.

Extra experiments were performed by connecting the vision network to a hand-tuned proportional-integral (PI) controller. All remaining details can be found in the supplementary materials.

\subsection*{Hardware setup}

Real-world experiments were performed with a custom-built quadrotor carrying the event-based camera (DAVIS240C), a single-board computer (UP Squared) and a neuromorphic processor (Intel Kapoho Bay with two Loihi neuromorphic research chips). A high-level overview can be found in \figref{fig:pipeline}, while all components are listed in \tabref{tab:components}. We use PX4\footnote{\url{https://px4.io/}} as autopilot firmware, and ROS\footnote{\url{https://ros.org/}} for communication. More specifically, events coming from the event-based camera are passed to the UP Squared over USB using ROS1. These events are processed (downsampling, cropping) on the UP Squared, and sent as spikes to the vision network running on the Kapoho Bay over USB. After processing, the output spikes are sent back over USB to the UP Squared, where they are decoded into the corner flows. The corner flows are then published by a ROS1 node, and sent to ROS2 over a ROS1-ROS2 bridge. The linear controller (or PI controller, for that matter) and the processing around it, running as a ROS2 node, takes the corner flows together with the attitude estimate coming from PX4 and the setpoint provided by the user, and outputs the control command. This command is then sent over ROS2 to PX4, and processed by the low-level controllers there. ROS2 makes use of RTPS for communication, which allows for high-frequency and high-bandwidth messaging between the UP and PX4, meaning our entire pipeline can run at 200~Hz. For position control between test runs, and as failsafe, we use an OptiTrack motion capture system.

\begin{table}[!h]
\centering
\caption{\textbf{List of hardware components used for the real-world test flights.}}
\label{tab:components}
\resizebox{0.475\textwidth}{!}{%
\begin{tabular}{@{}ll@{}}
\toprule
\textbf{Component}     & \textbf{Product}                \\ \midrule
Frame                  & GEPRC Mark 4 225~mm              \\ \midrule
Motor                  & Emax 2306 Eco II Series         \\ \midrule
Propellor              & Ethix S5 5~inch                 \\ \midrule
Battery                & Tattu FunFly 1800mAh 4S         \\ \midrule
Flight Controller      & Pixhawk 4 Mini                  \\ \midrule
ESC                    & SpeedyBee 45A BL32 4in1         \\ \midrule
Single-board computer  & UP Squared ATOM Quad Core 08/64 \\ \midrule
Event-based camera           & DAVIS240C                       \\ \midrule
Neuromorphic processor & Intel Loihi, Kapoho Bay form factor              \\ \bottomrule
\end{tabular}}
\end{table}

\subsection*{Energy benchmark}

Energy benchmarks for Loihi were performed on a Nahuku board\footnote{Host machine: Intel Xeon Platinum 8280 CPU at 2.7 GHz, 126 GB RAM, Ubuntu 20.04, NxSDK 1.0.0}, which contains 32 Loihi chips, as Kapoho Bay does not support energy probing. These (software) probes report a variety of power measurements, as well as execution times. Following the documentation, static power is due to transistor leakage, dynamic power is due to switching, idle power is measured while the embedded CPU cores on Loihi are still clocked but the neuromorphic are inactive, and total/running power is due to all components together. Kapoho Bay does support probing execution times, which we used to confirm that these are almost identical between Nahuku and Kapoho Bay. Furthermore, documentation states that Nahuku shuts down unused chips, meaning it can emulate energy consumption of a Kapoho Bay (which has two chips) with minimal overhead. Still, these benchmarks are not very representative of actual use on a drone, because that involves receiving and processing streaming data in an online fashion, whereas here all data is loaded to memory beforehand, and then processed as quickly as possible. Therefore, what these benchmarks represent is not the energy consumption and execution speed of the whole pipeline, but rather that of the network alone, without any bottlenecks and influences due to I/O and preprocessing. This also explains the much higher execution frequencies of Loihi with respect to real world tests, where this was always around 200 Inf/s.

On Jetson Nano, we simulate the SNN in PyTorch\footnote{ARM Cortex-A57 CPU at 1.43 GHz, 4 GB RAM, Ubuntu 20.04, PyTorch 1.12.0}. Jetson Nano has two power modes: a low-power (5W) mode, and a max-power (10W) mode, the difference being the number of active CPU cores (two versus four). Power consumption was measured using the tegrastats utility, while execution time was measured in Python code. Note that the split between static and dynamic power cannot be made here as these are not available as measurements. Idle power is measured for a period of time after running the benchmarks.

\section*{Acknowledgments}

This work was supported with funding from NWO (grants NWA.1292.19.298 and TOP grant 612.001.701), the Air Force Office of Scientific Research (award number FA8655-20-1-7044) and the Office of Naval Research Global (award number N629092112014). Furthermore, we are grateful to the Intel Neuromorphic Computing Lab and the Intel Neuromorphic Research Community for their support with Loihi.

\bibliography{sample}

\clearpage

\section*{Supplementary materials}

\subsection*{(More) Materials and methods}

\subsubsection*{Simulation}

Flightmare's quadrotor dynamics model~\cite{song2020flightmare} is as follows (notation from~\cite{kaufmann2022benchmark}):
\begin{align}
    \bm{\dot{p}}^\mathcal{WB} &= \bm{v}^\mathcal{WB} \\
    \bm{\dot{q}}^\mathcal{WB} &= \frac{1}{2}
    \begin{bmatrix}
        0 \\
        \bm{\omega}^\mathcal{B}
    \end{bmatrix}_\times
    \bm{q}^\mathcal{WB} \\
    \bm{\dot{v}}^\mathcal{WB} &= \frac{1}{m}\left(\bm{q}^\mathcal{WB} \odot \left(\bm{f}_\mathrm{prop} + \bm{f}_\mathrm{drag}\right)\right) + \bm{g}^\mathcal{W} \\
    \bm{\dot{\omega}}^\mathcal{B} &= \bm{J}^{-1}\left( \bm{\tau}_\mathrm{prop} - \bm{\omega}^\mathcal{B} \times \bm{J}\bm{\omega}^\mathcal{B} \right) \\
    \bm{\dot\Omega} &= \frac{1}{\tau_\Omega} \left( \bm{\Omega}_c - \bm{\Omega} \right) \label{eq:motors}
\end{align}
Here, $\bm{p}^\mathcal{WB}$ is the position of the body frame with respect to the world frame, $\bm{v}^\mathcal{WB}$ is the velocity of the body frame with respect to the world frame, $\bm{q}^\mathcal{WB}=(q_w,q_x,q_y,q_z)$ is a unit quaternion representing the orientation of the body frame with respect to the world frame, $[0,\bm{\omega}^\mathcal{B}]_\times^T \in \mathbb{R}^{4\times 4}$ is a skew-symmetric matrix representing infinitesimal rotations, $m$ is the drone's mass, $\odot$ is the quaternion-vector product, $\bm{f}_\mathrm{prop}$ is the collective thrust of all rotors along the body Z-axis, $\bm{f}_\mathrm{drag}$ is a drag force, $\bm{g}^\mathcal{W}$ is gravity in the world frame, $\bm{\omega}^\mathcal{B}$ is the angular rate of the body frame, $\bm{J} = \mathrm{diag}(J_x,J_y,J_z)$ is the drone's moment of inertia, $\bm{\tau}_\mathrm{prop}$ are the torques produced by the rotors, $\bm{\Omega}_{(c)}$ are the (commanded) motor speeds, and $\tau_\Omega$ is the time constant of the first order model of the motors.

The thrust and torque of each individual rotor contribute to the collective thrust and torque through:
\begin{align}
    \bm{f}_\mathrm{prop} &= \sum_i \bm{f}_i = m[0,0,\bar{f}]^T \\
    \bm{\tau}_\mathrm{prop} &= \sum_i \bm{\tau}_i + \bm{r}_i \times \bm{f}_i
\end{align}
where $\bar{f}$ is the mass-normalized thrust.

The thrust produced by a single rotor is modeled using a second-order thrust map, $\bm{f}_i=[0,0,a\Omega^2+b\Omega+c]^T$. Rotor drag is not modeled. Together with the motor dynamics in \eqnref{eq:motors}, this gives actual motor speeds, which give per-rotor thrusts and torques.

The outputs of the linear controller $\bm{c}\in \mathbb{R}^{4\times1}$ are then clamped to $[-1,1]$ and fed to different parts of the cascaded low-level controllers. First, $\bar{f}_{0,c}$ is scaled by a proportional gain $6g$ and set around hover: $\bar{f}_c = \bar{f}_{0,c}\cdot 6g + g$. Then, $\phi_c$ and $\theta_c$ are used by a proportional controller (gain of $\frac{\pi}{2}$) that sets the respective rates $\omega_{x,c}^\mathcal{B}$ and $\omega_{y,c}^\mathcal{B}$. Next, the three body rates are clamped and used by another proportional controller (respective gains of 16.6, 16.6, 5.0) to determine the desired body torques $\bm\tau_i$. The commanded clamped and mass-normalized thrust $\bar{f}_c$ is compensated for the attitude of the drone, partially dealing with the shortcomings of using a linear controller (explained in more detail in the next section).

All unmentioned constants and coefficients used in simulation can be found in \tabref{tab:simcoeffs}.

\begin{table}[!h]
\centering
\caption{\textbf{Coefficients used during simulation and training.}}
\label{tab:simcoeffs}
\resizebox{0.475\textwidth}{!}{%
\begin{tabular}{@{}lcc@{}}
\toprule
\textbf{Description}           & \textbf{Symbol}                                & \textbf{Value}                                                                        \\ \midrule
Drone mass                     & $m$ [kg]                                       & 1.535                                                                                 \\
Drone arm                      & $r$ [m]                                        & 0.255                                                                                 \\
Min. motor speed               & $\omega_\mathrm{min}$ [rpm]                    & 150                                                                                   \\
Max. motor speed               & $\omega_\mathrm{max}$ [rpm]                    & 1500                                                                                  \\
Motor time constant            & $\tau_\omega$ [s]                              & 0.025                                                                                 \\
Motor thrust map               & $a$, $b$, $c$                                    & \makecell{1.329825e-6,\\0.003836,\\-1.768999}                                                \\
Max. body rate                 & $\bm{\omega}^\mathcal{B}_\mathrm{max}$ [rad/s] & $[6.0, 6.0, 6.0]^T$                                                                   \\
Camera intrinsic matrix        & $\bm{K}$                                       & $\begin{bmatrix}188.84, 0, 90.0\\ 0, 188.99, 90.0\\ 0, 0, 1\end{bmatrix}$ \\
Camera rotation w.r.t. body    & $\bm{R}^\mathcal{CB}$                          & $\begin{bmatrix}1, 0, 0\\ 0, -1, 0\\ 0, 0, -1\end{bmatrix}$                           \\
Camera translation w.r.t. body & $\bm{T}^\mathcal{CB}$ [m]                      & $[-0.005, 0.077, -0.033]^T$                                                           \\
Lower altitude bound           & $p^\mathcal{WB}_{z,\mathrm{min}}$ [m]          & 0.2                                                                                   \\
\bottomrule
\end{tabular}}
\end{table}

\subsubsection*{Real world}

The vision network outputs corner flows in pixels per millisecond, while the user provides scaled velocity setpoints in 1/s. We connect these through manual scaling of $\bm{\hat{\nu}}^\mathcal{B}$ and of the control outputs. Furthermore, we smooth the computed visual observables, such that $\left[\bm{\hat{\nu}}^\mathcal{B},\hat{\omega}_z^\mathcal{B}\right]^T = \left[\bm{\hat{\nu}}^\mathcal{B},\hat{\omega}_z^\mathcal{B}\right]^T \circ \bm{\alpha}_{\nu,\omega} + (1 - \bm{\alpha}_{\nu,\omega}) \circ \bm{\beta}_{\nu,\omega} \circ \left[\bm{\hat{\nu}}^\mathcal{B},\hat{\omega}_z^\mathcal{B}\right]^T$.

During hover, we estimate hover thrust, and use this to offset $\bar{f}_{0,c}$. Because the drone's autopilot software uses a dimensionless number for thrust, we furthermore scale it by a proportional gain 0.3, and provide attitude compensation (discussed below) to limit the impact of using a linear controller. The other control commands are also scaled by a proportional gain of 0.3. During hover, we also calibrate attitude biases in roll and pitch, and use this to compensate both the inputs to the controller $\lvert\phi\rvert$ and $\lvert\theta\rvert$, as well as the control commands $\phi_c$ and $\theta_c$. While the control command in yaw is a rate, the autopilot software in attitude mode only accepts a yaw angle. To convert $\omega^\mathcal{B}_{z,c}$ to $\psi_c$, we use the yaw angle provided by an external motion capture system to initialize $\psi_c$, and then integrate using $\psi_c=\psi_c + 0.005\omega_{z,c}^\mathcal{B}$, with 0.005~s the approximate timestep of the loop.

Extra experiments were performed by connecting the vision network to a proportional-integral (PI) controller. Again, we smooth and scale (using the same constants) $\smash{\bm{\hat\nu}^\mathcal{B}}$ and $\smash{\hat{\omega}^\mathcal{B}_z}$, which are then compared against the setpoints $\smash{\bm{\nu}^\mathcal{B}_{sp}}$ and $\smash{\omega^\mathcal{B}_{z,sp}}$. The proportional and integral gains of the controller are hand-tuned and given in \tabref{tab:realcoeffs}, together with all constants used during the real-world tests.

\begin{figure*}[!t]
    \centering
    \includegraphics[width=\textwidth]{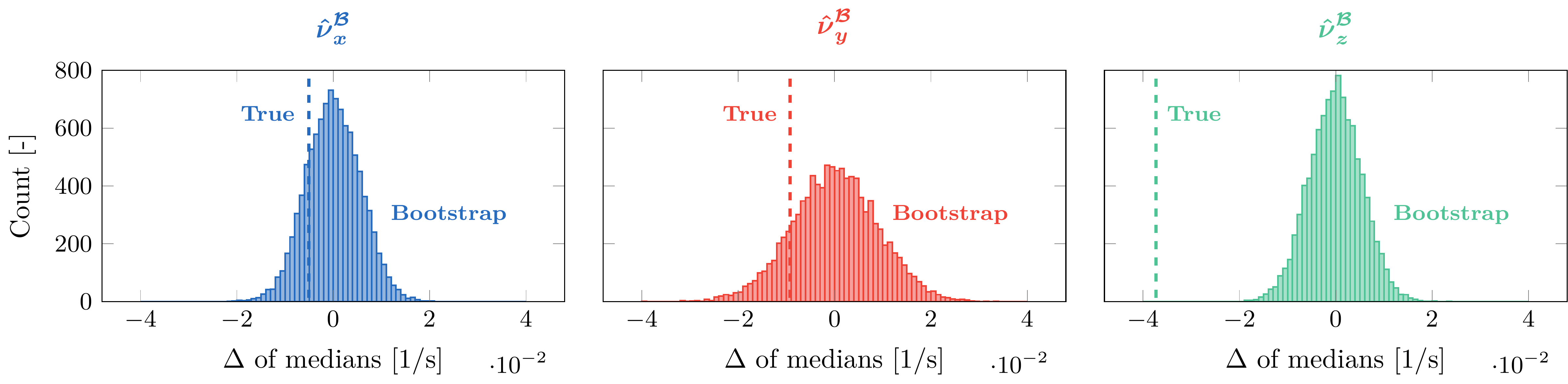}
    \caption{\textbf{Comparing flight tests with and without tether.} Real-world flight tests were performed with a tether. We perform ten runs with $\bm{{\nu}}^\mathcal{B}_{sp}=[0,0.5,0]^T$~1/s with and without this tether, and use bootstrapping to quantify its impact. More specifically, we combine all data points after the initial transient into a single set, sample two sets of 454 data points and compute the difference of their medians. We repeat this 10,000 times and compare it to the true difference of medians of the tests with and without tether. While the impact of the tether on $\hat{\nu}_z^\mathcal{B}$ is significant, it is also very small in terms of magnitude.}
    \label{fig:bootstrap}
\end{figure*}

As already hinted at before, a linear controller has several shortcomings that we have to be aware for a successful sim-to-real transfer. In short, these are:
\begin{itemize}
    \item A certain attitude angle is necessary to achieve a certain scaled velocity setpoint. As attitude changes, so does the fraction of the thrust vector that is pointing upward; this effect is nonlinear (small angle assumption not valid) and can therefore not be accounted for by the controller. Feeding absolute attitude angles allows the controller to do some compensation, but ad-hoc attitude compensation in the thrust is still necessary.
    \item The same scaled velocity value at different heights leads to different speeds, which needs different attitude angles to be achieved. This effect is also nonlinear, and hence we decided to carry out all horizontal flights from the same starting altitude. Note that this does not account for accumulated virtual drift during flight, and hence this effect will still be visible to some extent.
    \item When the drone is ascending or descending at a slope equal to its attitude, $\smash{\nu_z^\mathcal{B}=0}$ and hence additional information is necessary to counter this drift. Attitude of the drone can provide this information, but again this effect is nonlinear, and hence some drift will always remain.
\end{itemize}

These implications show different effects for different setpoints and between simulation and the real world because of their nonlinearity and because of the differences in characteristics between the simulated and real drone.

\begin{table}[!h]
\centering
\caption{\textbf{Coefficients used during real-world tests.}}
\label{tab:realcoeffs}
\resizebox{0.475\textwidth}{!}{%
\begin{tabular}{@{}lcc@{}}
\toprule
\textbf{Description}                          & \textbf{Symbol}            & \textbf{Value}               \\ \midrule
Visual observable smoothing                   & $\bm{\alpha}_{\nu,\omega}$ & $[0.90,0.90,0.95,0.90]^T$    \\
Visual observable scaling                     & $\bm{\beta}_{\nu,\omega}$  & $[0.9,0.9,1.0,1.0]^T$        \\
Proportional gains (thrust, roll, pitch, yaw) & $\bm{P}$                   & $[0.10,0.06,0.06,4.0]^T$     \\
Integral gains (thrust, roll, pitch, yaw)     & $\bm{I}$                   & $[0.0001,0.0003,0.0003,0]^T$ \\ \bottomrule
\end{tabular}}
\end{table}

\subsection*{Quantification of tether impact}

For safety reasons, real-world flight tests were performed with the drone attached to a lightweight tether. In \figref{fig:bootstrap} we quantify the impact of this tether through bootstrapping. We perform ten runs with $\bm{{\nu}}^\mathcal{B}_{sp}=[0,0.5,0]^T$~1/s with and without tether, and look at whether the true difference of medians lies close to the center of the distribution made up of the bootstrapped differences of medians. While this is the case for $\smash{\hat{\nu}^\mathcal{B}_x}$ and $\smash{\hat{\nu}^\mathcal{B}_y}$, this is not the case for $\smash{\hat{\nu}^\mathcal{B}_z}$. However, even though the impact of the tether in Z is significant, it is also very small. Furthermore, the Z-axis is most heavily influenced by external factors, such as proper hover thrust estimation, battery degradation, etc.

\end{document}